\documentclass[journal]{IEEEtran}
\usepackage{amsmath,amssymb,amsfonts}
\usepackage{graphicx}
\usepackage{placeins}
\usepackage{booktabs}
\usepackage{multirow}
\usepackage{cite}
\usepackage{bm}
\usepackage{xcolor}
\usepackage{url}
\usepackage[colorlinks=true,linkcolor=black,citecolor=black,urlcolor=blue]{hyperref}
\graphicspath{{figures/}}
\newcommand{\keyresult}[1]{\textbf{\boldmath #1}}
\newcommand{\ctau}{c_{\tau}}                        
\newcommand{\wcmd}[1][]{\omega_{\mathrm{cmd}#1}}    
\newcommand{\Lw}{L_{w}}                             
\newcommand{\Lin}{\mathcal{L}_{\mathrm{in}}}        

\begin{document}

\title{Actuator-Aware Observer Inputs for Cascade ADRC of Quadrotors Under Rotor Lag, Measurement Delay and Sensor Noise}

\author{
	Xuan-Quy~Dao\hspace{1mm}\href{https://orcid.org/0000-0001-5415-7538}
	{\includegraphics[scale=0.06]{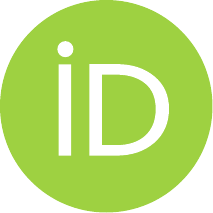}},
	Ngoc-Bich~Le\hspace{1mm}\href{https://orcid.org/0000-0001-7431-0157}
	{\includegraphics[scale=0.06]{orcid.pdf}},
	and
	Le-Minh~Nguyen\hspace{1mm}\href{https://orcid.org/0009-0002-0384-3938}
	{\includegraphics[scale=0.06]{orcid.pdf}}

	\thanks{X.-Q. Dao and L.-M. Nguyen are with Eastern International University,
		Ho Chi Minh City, Vietnam
		(e-mail: \texttt{quy.dao@eiu.edu.vn};
		\texttt{minh.nguyenle@eiu.edu.vn}).}%
	\thanks{N.-B. Le is with Industrial University of Ho Chi Minh City,
		Ho Chi Minh City, Vietnam
		(e-mail: \texttt{lengocbich@iuh.edu.vn}).}
}

\markboth{}{Dao \MakeLowercase{\textit{et al.}}:
	Actuator-Aware Observer Inputs for Cascade ADRC of Quadrotors}

\maketitle

\begin{abstract}
In sampled cascade linear active disturbance rejection control (LADRC) of quadrotors, the extended-state observer is usually driven by the instantaneous command, although the delayed measurement reflects an earlier, rotor-filtered acceleration. This simulation and software-in-the-loop (SITL) study examines actuator-aware (AA) observer inputs: the altitude and attitude observers use the limited virtual input filtered by a rotor-lag model and aligned with the measurement delay, and the horizontal observers use the acceleration reconstructed from the measured attitude. Fixed gains are selected on the sampled model under sensitivity and step-response requirements and covariance-based ceilings on the noise-induced control increment, including the path from position noise to the inner loop. In 30 paired set-point trials, the two AA designs reach position root-mean-square errors (RMSEs) of 19.2 and 16.9~mm, versus 65.5~mm for a proportional-integral-derivative (PID) controller tuned under the same requirements, with lower rotor-command increments. Ablations attribute the improvement to the rotor-lag model and the measured-attitude input; delay alignment is negligible at the nominal 5-ms delay. PX4 SITL runs reproduce these results within 2~mm. Behind the PX4 state estimator, the white-noise ceilings no longer describe the loop. After redesign with an identified estimation-error model, one AA design reduces the set-point RMSE from 59.1 to 44.0~mm with a 15\% lower quantized rotor-command increment, whereas the half-gain variant is worse than the PID. With fixed gains, the AA designs tolerate less unmodelled rotor lag and delay than the PID. Hardware validation remains future work.
\end{abstract}


\section{Introduction}
Quadrotor trajectory control couples disturbance rejection, sensor-noise sensitivity, actuator dynamics, and measurement timing \cite{mahony2012}. Active disturbance rejection control (ADRC) estimates a lumped disturbance using an extended-state observer (ESO) and compensates for it in the control law \cite{han2009,gao2003}. In a sampled cascade, however, the instantaneous commanded input may differ from the acceleration associated with a delayed measurement: rotor dynamics filter the command, and the measurement refers to an earlier plant state. Increasing observer bandwidth can then amplify both measurement noise and the residual caused by this input inconsistency.

\begin{figure*}[t]
	\centering
	\includegraphics[width=\textwidth]{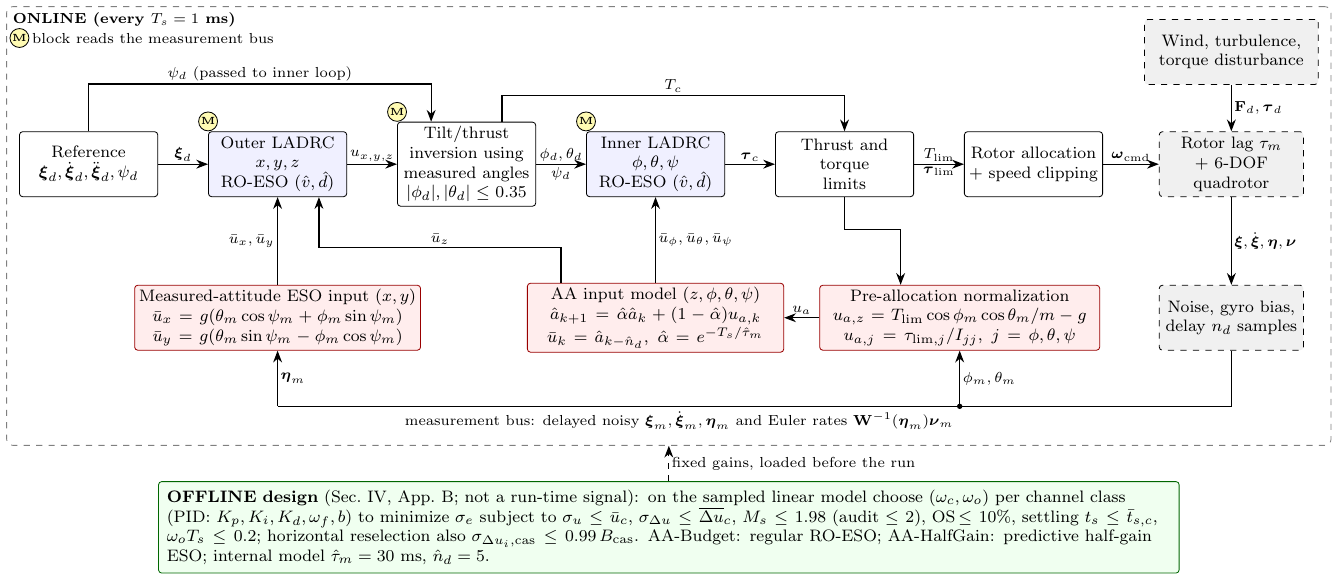}
	\caption{Actuator-aware cascade LADRC. The dashed frame is the online loop executed at every $T_s$; blocks marked M read the delayed noisy measurements. The red blocks are the two observer-input paths: measured attitude for $x,y$, and the delay-aligned actuator model for $z,\phi,\theta,\psi$. The gains are fixed offline.}
	\label{fig:architecture}
\end{figure*}

This work addresses that implementation problem with two distinct actuator-aware (AA) observer-input paths (Fig.~\ref{fig:architecture}). For altitude and attitude, the controller reconstructs normalized inputs from the limited thrust and torque before rotor allocation, propagates them through a first-order actuator model, and aligns the result with the measurement time. For horizontal motion, the observer uses near-hover acceleration reconstructed from measured attitude rather than the instantaneous tilt command. The nominal model uses unity normalized input gain and separates the actual plant lag and delay from the internal observer assumptions. These choices do not reconstruct post-allocation rotor clipping or adapt the assumed delay during variable-latency tests.

Observer-input consistency is combined with fixed-gain activity-aware tuning. The sampled linear covariance model includes the closed attitude loop, its observer, actuator lag, and measurement buffers. It predicts tracking error, noise-induced virtual-control variance, and sample-to-sample increments. During horizontal gain selection, an additional inner-control increment ceiling includes the propagation of position and velocity noise through tilt commands. Sensitivity, overshoot and settling-time requirements constrain the same candidate search; the design target is $M_s\le1.98$, followed by a separate final audit against $M_s\le2$.

Actuator models in observers, delay alignment and half-gain tuning each have precedents; the contribution lies in two design concepts. The first is \emph{observer-input consistency} in a sampled cascade with rotor lag and measurement delay. A commanded observer input adds the structural residual \eqref{eq:input_mismatch} to the disturbance estimate, so the sensitivity peak rises with observer bandwidth at fixed controller bandwidth (Section~\ref{sec:nominal}). In the redesigned ablation, the response and sensitivity requirements are then met only at the lowest observer/controller ratios of the searched grid; there, the best commanded-input design found for $\omega_o/\omega_c\ge1$ has 9.2 times the mean root-mean-square error (RMSE) of AA-Budget (Section~\ref{sec:abl}). This ratio describes the feasible designs of the stated procedure rather than an input swap at fixed gains; the fixed-bandwidth mechanism is shown by the linear analysis of Fig.~\ref{fig:msratio}. The second is that the \emph{outer-to-inner noise path} becomes an explicit design constraint. The cascade-level increment $\sigma_{\Delta u_i,\mathrm{cas}}$ is predicted on the full cascade model and agrees with nonlinear hover measurements, which permits a comparison under a common strict cascade-increment ceiling (Section~\ref{sec:cascade}). With a shaping filter for the estimation error, the same constraint is computed for estimator-fed loops (Section~\ref{sec:estaware}).

The main contributions are:
\begin{enumerate}
\item Two actuator-aware observer-input paths for sampled cascade linear ADRC (LADRC), actuator-model inputs aligned with the measurement time (altitude/attitude) and measured-attitude inputs (horizontal). Ablations show that the improvement comes from the rotor-lag model and the measured-attitude input, whereas the alignment itself has a negligible effect at the nominal 5-ms delay.
\item A fixed-gain covariance design whose channel-level and cascade-level activity ceilings are computed for white or estimator-coloured measurement noise, validated on code-generated controllers inside PX4.
\item A paired nonlinear simulation assessment, including a comparison under a common strict cascade-increment ceiling, noise and latency stress, and fixed-gain rotor-lag and delay mismatch, under which the AA designs are less tolerant than proportional-integral-derivative (PID) control in the linear sensitivity screen.
\item A repetition in PX4 software-in-the-loop with the controllers code-generated into the autopilot, including a redesign for the estimator-fed loop under the same sensitivity and response specification and estimator-aware increment ceilings.
\end{enumerate}

Section~\ref{sec:related} reviews related work, and Section~\ref{sec:model} states the quadrotor model and measurement assumptions. Section~\ref{sec:proposed} presents the actuator-aware observer inputs and the activity-constrained tuning. Section~\ref{sec:results} reports the simulation and PX4 software-in-the-loop results, Section~\ref{sec:discussion} discusses scope and limitations, and Section~\ref{sec:conclusion} concludes.

\section{Related Work}\label{sec:related}

\emph{Observer-based robust control.}
Disturbance-observer and adaptive-control methods for quadrotor disturbance rejection are reviewed in \cite{wang2024survey}. Within the ADRC/ESO family, representative developments include cascade LADRC position control with PID attitude stabilization \cite{ramirez2024}, generalized proportional-integral (GPI) observer-based ADRC for morphing quadrotors \cite{he2023gpi}, reduced-order ESO-based nonlinear PID control \cite{xu2025roeso}, and geometric ADRC on $\mathrm{SO}(3)$ \cite{zeng2025so3}. Related nonlinear-observer approaches include fixed-time extended observers combined with adaptive sliding-mode control under turbulent wind \cite{miranda2023fxeso}, fixed-time disturbance observers with backstepping under input saturation and actuator failure \cite{huang2025fxtbs}, high-order sliding-mode disturbance observers incorporating first-order-plus-dead-time actuator dynamics \cite{xu2024hosmdo}, fixed-time disturbance-observer-based model predictive control (MPC) \cite{xu2024fxtdompc}, high-gain disturbance observers \cite{izadi2023hgdo}, and neural-network disturbance observers with command filtering \cite{li2025nndo}. Their analyses hold under their stated modeling and disturbance assumptions and do not directly cover Gaussian sensor noise, measurement-timing effects on the observer input, or rotor-lag mismatch in a sampled nonlinear cascade.

\emph{Tuning of ADRC and PID.}
Bandwidth parameterization reduces LADRC tuning to $(\omega_c,\omega_o)$ \cite{gao2003}, and stability with uncertain dynamics was studied in \cite{zheng2007}. Quadrotor studies automate tuning with Kriging surrogates \cite{zhou2024kriging}, crayfish optimization with fuzzy adaptation \cite{li2025ladrccoa}, particle-swarm tuning with event triggering \cite{li2025etftadrc}, twin-delayed deep deterministic policy gradient (TD3) reinforcement learning \cite{cai2026td3adrc}, and multi-objective PID optimization \cite{vaiuso2025mopid}. Their costs emphasize tracking and transients, in some cases with input-smoothness, power or acoustic terms \cite{cai2026td3adrc,vaiuso2025mopid}; these objectives differ from the stationary sensor-noise covariance constraints on virtual-control variance, increments and the outer-to-inner noise path used here.

\emph{Noise-aware observers.}
The trade-off between observer responsiveness and measurement-noise sensitivity has been addressed through error-dependent observer bandwidth adaptation \cite{liu2020beso,cui2025vbeso}, half-gain tuning of discrete-time ESOs \cite{herbst2025halfgain}, and scalable-bandwidth linear ESOs combined with internal model control for quadrotor attitude regulation \cite{wei_scalable_leso}. These approaches modify the observer bandwidth or gain structure; the present work instead specifies the input and noise paths of a sampled cascade model and combines them with fixed-gain covariance constraints (Section~\ref{sec:proposed}). Two studies are closest in ingredients. Zhang et al.~\cite{zhang2022meso} place an actuator model inside the ESO, and Herbst and Hempel~\cite{herbst2025halfgain} derive half-gain tuning for discrete-time ESOs and validate it experimentally. The present work uses both ingredients and claims neither as new: the actuator-aware input is related to the model-based ESO, and the half-gain observer is adapted from~\cite{herbst2025halfgain}. Neither study addresses what differs here: the measured-attitude input of the horizontal observers and the selection of fixed gains under cascade-level and estimator-aware increment ceilings. Reconstructing horizontal acceleration from the realized attitude is common near-hover practice; what is examined here is its use as the input of a reduced-order ESO under these ceilings. Their reported results are not directly comparable, because plants and objectives differ. Table~\ref{tab:lit} summarizes selected features.

\emph{Delay compensation and input synchronization.} Matching the observer input to the time of the measurement is not new. Predictive ADRC~\cite{zheng2014padrc} makes the ESO predictive for processes with a known delay, modified ADRC~\cite{zhao2014madrc} modifies ADRC for time-delay systems, and~\cite{skupin2022delay} synchronizes the observer input with a process delay. Incremental nonlinear dynamic inversion (INDI) for multirotors~\cite{smeur2016aindi,tal2021indi} is structurally close to \eqref{eq:aa}: it combines the actuator state, modelled~\cite{smeur2016aindi} or measured with motor encoders~\cite{tal2021indi}, with an angular-acceleration estimate on a common time base and inverts the control effectiveness incrementally. INDI therefore needs an angular-acceleration estimate from differentiated rate measurements, whereas the ESO estimates the lumped disturbance from the delayed angle or position. The measurement-time alignment of \eqref{eq:aa} is used here as a known ingredient and is isolated by ablation A2 (Section~\ref{sec:abl}). Among the reported designs in Table~\ref{tab:lit}, only adaptive INDI~\cite{smeur2016aindi} combines an actuator model with a timing correction, by synchronizing the delay of its angular-acceleration estimate with the actuator feedback; it does not treat a delayed position or attitude measurement that drives an observer. None of the listed designs constrains the control increment or the cascade noise path during tuning.

\begin{table*}[t!]
	\centering
	\scriptsize
	\caption{Selected Features of Related Methods}
	\label{tab:lit}
	\setlength{\tabcolsep}{2.3pt}
	\renewcommand{\arraystretch}{1.10}
	\begin{tabular}{llcccccll}
		\toprule
		\textbf{Study} &
		\textbf{Method} &
		\textbf{Noise} &
		\shortstack{\textbf{Actuator dynamics}} &
		\textbf{Delay} &
		\shortstack{\boldmath$\mathbf{\Delta u}$ limit} &
		\shortstack{\textbf{Cascade path}} &
		\textbf{Tuning/design} &
		\textbf{Validation} \\
		\midrule
		
		\cite{ramirez2024}
		& LADRC/GPI cascade
		& -- & -- & -- & -- & --
		& observer poles
		& flight \\
		
		\cite{zhou2024kriging}
		& ADRC
		& -- & -- & -- & -- & --
		& Kriging surrogate
		& HIL \\
		
		\cite{li2025ladrccoa}
		& LADRC/fuzzy
		& -- & -- & -- & -- & --
		& modified crayfish
		& simulation \\
		
		\cite{cai2026td3adrc}
		& TD3--ADRC
		& -- & sim.$^{a}$ & -- & --$^{b}$ & --
		& TD3
		& bench \\
		
		\cite{zhang2022meso}
		& model-based ESO
		& -- & yes & -- & -- & --
		& actuator-model ESO
		& simulation \\
		
		\cite{smeur2016aindi}
		& adaptive INDI
		& -- & yes & sync.$^{f}$ & -- & --
		& online effectiveness
		& flight \\
		
		\cite{tal2021indi}
		& INDI/flatness
		& -- & meas.$^{g}$ & -- & -- & --
		& INDI, flatness
		& flight \\
		
		\cite{zheng2014padrc}
		& predictive ADRC
		& -- & -- & yes & -- & --
		& predictive ESO
		& sim., HIL \\
		
		\cite{skupin2022delay}
		& RADRC/RESO
		& -- & -- & yes$^{e}$ & -- & --
		& delay synchronization
		& analysis \\
		
		\cite{bai2023pde}
		& predictor/cascade ESO
		& yes & -- & yes & -- & --
		& PDE predictor/ESO
		& experiment \\
		
		\cite{xu2024hosmdo}
		& HOSM observer
		& -- & yes$^{c}$ & -- & -- & --
		& stability conditions
		& simulation \\
		
		\cite{wei_scalable_leso}
		& LESO/IMC
		& yes & -- & -- & -- & --
		& scalable bandwidth
		& HIL \\
		
		\cite{cui2025vbeso}
		& variable-bandwidth ESO
		& yes & -- & -- & -- & --
		& error-adaptive bandwidth
		& spacecraft HIL \\
		
		\cite{michalski2024kfeso}
		& KF-tuned ESO
		& yes & -- & -- & -- & --
		& noise-variance tuning
		& laboratory \\
		
		\cite{herbst2025halfgain}
		& half-gain ESO
		& yes & -- & -- & -- & --
		& analytic half-gain
		& experiment \\
		
		\textbf{This work}
		& \textbf{AA cascade LADRC}
		& \textbf{yes}$^{d}$
		& \textbf{yes}
		& \textbf{yes}
		& \textbf{local+cascade}
		& \textbf{yes}
		& \textbf{constrained search}
		& \textbf{sim., PX4 SITL} \\
		
		\bottomrule
	\end{tabular}
	
	\par\footnotesize
	``Noise'' denotes sensor noise explicitly considered in observer or
	tuning design. Actuator dynamics included in the
	design model. ``Delay'' denotes output or measurement delay and excludes
	delay internal to an actuator model. ``Cascade path'' denotes explicit
	propagation of outer-loop measurement noise through tilt commands into
	inner-loop control activity.
	$^{a}$Rotor lag is included as a simulation condition rather than as an
	explicit activity constraint in tuning.
	$^{b}$A control-magnitude term in the reward is not a sample-to-sample
	control-increment constraint.
	$^{c}$The FOPTD actuator model includes actuator time delay; this is
	classified under actuator dynamics rather than output/measurement delay.
	$^{d}$White measurement noise and estimator-coloured noise identified in PX4 SITL.
	$^{e}$Process (dead-time) delay of a first-order-plus-dead-time plant, equivalent to an output delay for the single-input single-output plant.
	$^{f}$Delay of the angular-acceleration estimate synchronized with the actuator feedback.
	$^{g}$Motor speed measured with encoders and controlled in closed loop.
	HIL: hardware-in-the-loop; SITL: software-in-the-loop; INDI: incremental nonlinear dynamic inversion; RADRC/RESO: reduced-order ADRC/ESO; PDE: partial differential equation; HOSM: high-order sliding mode; LESO: linear ESO; IMC: internal model control; KF: Kalman filter; FOPTD: first-order plus time delay.
	Here, ``--'' denotes a feature not identified in the reported design; this
	does not establish that the feature is absent from the cited method. Because the studies differ in plants, objectives and
	conditions, the table is a feature-level comparison, not a performance ranking.
\end{table*}

\section{Quadrotor Model and Measurement Assumptions}\label{sec:model}

The plant is a ``+''-configuration quadrotor with the nominal parameters of \cite{luukkonen2011} (Table~\ref{tab:params}). Let $\boldsymbol{\xi}\in\mathbb{R}^3$ denote the inertial position, $\boldsymbol{\eta}=[\phi,\theta,\psi]^{\top}$ the ZYX Euler angles, $\boldsymbol{\nu}\in\mathbb{R}^3$ the body angular velocity, and $\omega_i$ the speed of rotor $i$. The total thrust and body torques are
\begin{equation}
	T=k\sum_{i=1}^{4}\omega_i^2,\qquad
	\boldsymbol{\tau}=
	\begin{bmatrix}
		lk(\omega_4^2-\omega_2^2)\\
		lk(\omega_3^2-\omega_1^2)\\
		\ctau(-\omega_1^2+\omega_2^2-\omega_3^2+\omega_4^2)
	\end{bmatrix},
	\label{eq:mix}
\end{equation}
and the Newton--Euler dynamics are \cite{luukkonen2011,mahony2012}
\begin{align}
	m\ddot{\boldsymbol{\xi}}
	&=
	-mg\,\mathbf{e}_3
	+\mathbf{R}(\boldsymbol{\eta})T\mathbf{e}_3
	-\mathbf{A}(\dot{\boldsymbol{\xi}}-\mathbf{v}_w)
	+\mathbf{F}_d,
	\label{eq:trans}\\
	\mathbf{I}\dot{\boldsymbol{\nu}}
	&=
	-\boldsymbol{\nu}\times(\mathbf{I}\boldsymbol{\nu})
	-I_M(\boldsymbol{\nu}\times\mathbf{e}_3)\Omega_r
	+\boldsymbol{\tau}
	+\boldsymbol{\tau}_d,
	\label{eq:rot}\\
	\dot{\boldsymbol{\eta}}
	&=
	\mathbf{W}^{-1}(\boldsymbol{\eta})\boldsymbol{\nu}.
	\label{eq:kin}
\end{align}
Here $\mathbf e_3=[0,0,1]^\top$, $\mathbf R=R_z(\psi)R_y(\theta)R_x(\phi)$ maps body to inertial coordinates, $\mathbf I=\operatorname{diag}(I_{xx},I_{yy},I_{zz})$, $\mathbf A=A\mathbf I_3$, and $\mathbf W^{-1}(\boldsymbol\eta)$ is the ZYX Euler-rate map, defined for $\cos\theta\ne0$ \cite{luukkonen2011}. The wind velocity is $\mathbf{v}_w$; $\mathbf F_d$ (inertial) and $\boldsymbol\tau_d$ (body) are external force and torque disturbances, and $\Omega_r=\omega_1-\omega_2+\omega_3-\omega_4$. The scalar $k$ in \eqref{eq:mix} is the thrust coefficient; subscripts $k$ elsewhere denote the sample index.

Unlike an ideal instantaneous actuator, each rotor is modeled by a
first-order response from commanded to actual speed
\cite{powers2015},
\begin{equation}
	\tau_m\dot{\omega}_i
	=
	\wcmd[,i]-\omega_i,
	\qquad
	0\leq\wcmd[,i]\leq\omega_{\max},
	\label{eq:motor}
\end{equation}
where $\tau_m$ is the rotor time constant. The commands $\wcmd[,i]$
are obtained by inverting \eqref{eq:mix} for the requested
$(T_c,\boldsymbol{\tau}_c)$, followed by rotor-speed saturation.

Linearizing about symmetric hover, $\omega_i=\omega_h+\delta\omega_i$, gives $\delta T=2k\omega_h\sum_i\delta\omega_i$ and
\begin{equation}
	\tau_m\dot{\delta T}=\delta T_c-\delta T,
	\label{eq:thrustlag}
\end{equation}
with the same time constant for the small-signal torques. The first-order lag used below for the virtual thrust and attitude inputs is therefore a local approximation of the rotor-speed dynamics.

\begin{table}[t]
\centering
\caption{Nominal Model and Simulation Parameters}\label{tab:params}
\footnotesize\setlength{\tabcolsep}{2pt}
\begin{tabular}{@{}llr@{}}
\toprule
Parameter & Symbol & Value\\
\midrule
Mass, arm length$^{a}$ & $m$, $l$ & 0.468 kg, 0.225 m\\
Thrust coeff.$^{a}$ & $k$ & $2.98{\times}10^{-6}$ N\,s$^2$\\
Drag-torque coeff.$^{a}$ & $\ctau$ & $1.14{\times}10^{-7}$ N\,m\,s$^2$\\
Inertia$^{a}$ & $I_{xx}{=}I_{yy}$, $I_{zz}$ & $(4.856,8.801){\times}10^{-3}$ kg\,m$^2$\\
Rotor inertia, linear drag$^{a}$ & $I_M$, $A$ & $3.357{\times}10^{-5}$ kg\,m$^2$, 0.25 kg/s\\
Rotor time constant$^{c}$ & $\tau_m$ & 30 ms\\
Measurement delay$^{c}$ & $n_d T_s$ & 5 ms ($n_d=5$, $T_s=1$ ms)\\
Rotor / thrust limits$^{c}$ & $\omega_{\max}$, $T_{\max}$ & 1000 rad/s, $2mg$\\
Torque limits$^{c}$ & $\tau_{\max}$ & (0.2, 0.2, 0.05) N\,m\\
Noise std (pos., vel.)$^{c}$ & $\sigma_p,\sigma_v$ & 0.01 m, 0.02 m/s\\
Noise std (angle, gyro)$^{c}$ & $\sigma_\eta,\sigma_g$ & 0.0035 rad, 0.005 rad/s\\
Gyro-bias random walk$^{c}$ & $\sigma_b$ & $10^{-4}$ rad/s/$\sqrt{\text{s}}$\\
Mean wind$^{c}$ & $\bar{\mathbf v}_w$ & (1, 0.5, 0) m/s\\
Turbulence$^{c}$ & $\sigma_w$, $\Lw/V$ & 0.5 m/s ($z$: 0.25), 1 s\\
\bottomrule
\end{tabular}\\[2pt]
\footnotesize $^{a}$\cite{luukkonen2011}; $^{c}$assumed in this study.
\end{table}

\subsection{Sensors, delay and wind}
Position, velocity, Euler angles, and body rates carry additive zero-mean Gaussian white noise, and the gyroscope also has a random-walk bias, following the stochastic sensor-error models of \cite{gallo2022sensors}. The controller reconstructs Euler rates as $\mathbf{W}^{-1}(\boldsymbol{\eta}_m)\boldsymbol{\nu}_m$; the independent Euler-rate noise of the hover covariance model is a local approximation of this transformation. All measurements, including the gyro bias, reach the controller after a common delay of $n_d$ samples that represents sensing, estimation, and computation latency; the history is initialized with the first measurement, and $n_{\chi,k}$ in \eqref{eq:chan} denotes the noise of the delayed sample available at $k$. Wind is a mean component plus independent first-order Gauss--Markov turbulence per inertial axis with correlation time $\tau_w=\Lw/V$, where $\Lw/V$ is the scale-length-to-airspeed time scale of the Dryden spectrum \cite{zhou2022gust}. The first-order form is a colored-disturbance surrogate applied to every axis, not a full Dryden or von K\'arm\'an model. It enters through the drag term in \eqref{eq:trans}.

\subsection{Normalized channel model}
As is standard in quadrotor cascade control
\cite{mahony2012}, the outer loop generates the horizontal
acceleration commands $u_x$ and $u_y$ and the vertical acceleration command
$u_z$, while the inner loop regulates the attitude
$(\phi,\theta,\psi)$. All controllers use the same command-to-force/attitude
mapping. The requested thrust and body torque are
\begin{align}
	T_c &=
	\frac{m(g+u_z)}{\cos\phi_m\cos\theta_m},
	&
	\boldsymbol{\tau}_c
	&=
	\mathbf{I}
	\begin{bmatrix}
		u_\phi & u_\theta & u_\psi
	\end{bmatrix}^{\top},
	\label{eq:inv1}
\end{align}
while the horizontal acceleration commands are converted to roll and pitch
references using the near-hover small-angle inversion
\begin{equation}
	\theta_d=\tfrac{u_x\cos\psi_m+u_y\sin\psi_m}{g},\;
	\phi_d=\tfrac{u_x\sin\psi_m-u_y\cos\psi_m}{g}.
	\label{eq:inv2}
\end{equation}
The right-hand sides in \eqref{eq:inv2} are clipped componentwise to $[-0.35,0.35]$ rad. All angles in the implemented inversion are measured angles. Before rotor allocation, thrust and torque are limited as
$T_{\rm lim}=\operatorname{clip}(T_c,0,T_{\max})$ and
$\tau_{{\rm lim},j}=\operatorname{clip}(\tau_{c,j},-\tau_{\max,j},\tau_{\max,j})$,
where $\operatorname{clip}(a,l,u)=\min\{\max\{a,l\},u\}$. The actuator-aware inputs are
\begin{equation}
 u_{a,z}=\frac{T_{\rm lim}\cos\phi_m\cos\theta_m}{m}-g,\quad
 u_{a,j}=\frac{\tau_{{\rm lim},j}}{I_{jj}},
 \label{eq:limited_input}
\end{equation}
for $j\in\{\phi,\theta,\psi\}$, with $I_{\phi\phi}=I_{xx}$, $I_{\theta\theta}=I_{yy}$, and $I_{\psi\psi}=I_{zz}$. This normalization neglects Euler/body-rate coupling away from hover. Rotor allocation and individual-rotor clipping follow these componentwise limits.

For observer design and bandwidth selection, each controlled coordinate
$\chi\in\{x,y,z,\phi,\theta,\psi\}$ is represented by the normalized
second-order channel
\begin{equation}
	\ddot{\chi}=d_\chi+a_\chi ,
	\label{eq:chan_cont}
\end{equation}
where $d_\chi$ collects the unmodeled dynamics, coupling terms, and external
disturbances, and $a_\chi$ denotes the acceleration generated by the
actuator-mediated control input. At the controller sampling period $T_s$,
the measurement and actuator dynamics are represented as
\begin{equation}
	\begin{aligned}
		\chi_{m,k}
		&=
		\chi_{k-n_d}+n_{\chi,k},\\
		a_{\chi,k+1}
		&=
		\alpha a_{\chi,k}
		+(1-\alpha)b_\chi u_{\chi,k},
		\qquad
		\alpha=e^{-T_s/\tau_m},
	\end{aligned}
	\label{eq:chan}
\end{equation}
where $b_\chi$ is the normalized input-gain ratio. The actuator recursion in \eqref{eq:chan} applies to $\chi\in\{z,\phi,\theta,\psi\}$; the horizontal channels use the embedded attitude dynamics described below. In the nominal
covariance and actuator-aware design models, $b_\chi$ is set to unity; residual
input-gain error is absorbed into $d_\chi$. The corresponding rate measurement
$\dot{\chi}_{m,k}$ is subject to the same measurement delay.

For the horizontal channels, $a_x$ and $a_y$ are not produced directly by
the rotors but through the closed attitude loop. Under the same near-hover
approximation used in \eqref{eq:inv2},
\begin{align}
	a_x &\approx
	g(\theta\cos\psi+\phi\sin\psi),\\
	a_y &\approx
	g(\theta\sin\psi-\phi\cos\psi).
\end{align}
Hence, the outer-loop plant explicitly contains the inner-loop attitude
response, which is the mechanism by which position-measurement noise can
propagate through the commanded tilt into inner-loop control activity.

\section{Proposed Actuator-Aware Cascade LADRC and Activity-Constrained Tuning}\label{sec:proposed}
\subsection{Reduced-order LADRC}
Figure~\ref{fig:architecture} summarizes the cascade architecture and the two actuator-aware observer-input paths. The dashed frame is the online loop executed at every $T_s$, and the blocks marked M read the delayed noisy measurements. The red blocks are the two observer-input paths: measured attitude for $x,y$ and the delay-aligned actuator model for $z,\phi,\theta,\psi$. The gains are fixed offline before the run. AA-Budget and AA-HalfGain share these input paths and differ in the observer gain structure described below.

Each LADRC channel employs a rate-aided reduced-order extended-state
observer (RESO) to estimate the channel rate and lumped disturbance
\cite{han2009,gao2003,xu2025roeso}:
\begin{align}
	\hat v_{k+1}
	&=
	\hat v_k
	+T_s\!\left[
	\hat d_k+\bar u_k
	+\beta_1\!\left(\dot{\chi}_{m,k}-\hat v_k\right)
	\right],
	\nonumber\\
	\hat d_{k+1}
	&=
	\hat d_k
	+T_s\beta_2\!\left(\dot{\chi}_{m,k}-\hat v_k\right),
	\qquad
	\beta_1=2\omega_o,\quad
	\beta_2=\omega_o^2 ,
	\label{eq:reso}
\end{align}
where $\bar u_k$ is the input supplied to the observer model. The corresponding
LADRC law is
\begin{equation}
	u_k =
	\omega_c^2(\chi_{d,k}-\chi_{m,k})
	+2\omega_c(\dot{\chi}_{d,k}-\hat v_k)
	+\ddot{\chi}_{d,k}
	-\hat d_k .
	\label{eq:law}
\end{equation}
Here, $\omega_c$ and $\omega_o$ denote the controller and observer
bandwidths, respectively. The derivative-reference and acceleration-feedforward terms in \eqref{eq:law} are enabled for translational channels. The implemented attitude loops set both reference-derivative inputs to zero, including for time-varying commanded tilt. For fixed gains, the commanded-input and actuator-aware formulations differ only in the construction of $\bar u_k$.

\subsection{PID baseline}\label{sec:pid}
The PID baselines use a two-degree-of-freedom law with set-point weighting~\cite{astrom2006} in the same channel coordinates, read the same delayed noisy measurement bus as the LADRC channels (no delay compensation), and share the command inversion, limits and mixer. With $\chi_{d,k}$ the reference and $\dot\chi_{m,k}$ the measured rate (for attitude, the Euler rates $\mathbf W^{-1}(\boldsymbol\eta_m)\boldsymbol\nu_m$), each channel evaluates at sample $k$
\begin{equation}
	u_k=K_p\left(b\,\chi_{d,k}-\chi_{m,k}\right)+K_i\,\iota_k+K_d\left(\dot\chi_{d,k}-v_{f,k}\right)+\ddot\chi_{d,k},
	\label{eq:pid}
\end{equation}
and then updates its states with the same sample:
\begin{align}
	\iota_{k+1}&=\operatorname{sat}_{\ell_k}\!\left[\iota_k+T_s\left(\chi_{d,k}-\chi_{m,k}\right)\right],\nonumber\\
	\ell_k&=\left(i_{\lim}+K_p(1-b)\,|\chi_{d,k}|\right)/K_i,\nonumber\\
	v_{f,k+1}&=a_f v_{f,k}+(1-a_f)\,\dot\chi_{m,k},\qquad a_f=e^{-\omega_fT_s},
	\label{eq:pidstate}
\end{align}
where $\operatorname{sat}_{\ell}$ clips to $[-\ell,\ell]$, $i_{\lim}=(2,3,20,5)$ for $(xy,z,\phi\theta,\psi)$, and $\iota_0=v_{f,0}=0$: all controller states start at zero, and in PX4 they are reset while the vehicle is disarmed. The integrator stores the integrated unweighted error, and the derivative action uses the filtered rate of the previous samples. Because the proportional term acts only on $b\chi_{d}$, the integral action $K_i\iota$ must supply $K_p(1-b)\chi_d$ in steady state; the clamp on $K_i\iota$ is widened by exactly this share plus the disturbance allowance $i_{\lim}$. The clamp is the only anti-windup mechanism: the integrator is not frozen when the tilt clipping of \eqref{eq:inv2} or the thrust and torque limits before \eqref{eq:limited_input} are active. As for LADRC, the reference-derivative and acceleration terms are used in the translational channels and set to zero in the attitude channels. PID-Budget, PID-Budget-S, PID-E and the PID points of Fig.~\ref{fig:budget} use this law with different parameters $(K_p,K_i,K_d,\omega_f,b)$; the same function is called in the MATLAB simulation, the Simulink model and the code generated for PX4, and a sample-by-sample test reproduces the implementation exactly (\texttt{pid\_equation\_check.m}, output \texttt{pid\_equation\_consistency.csv}).

\emph{Remarks on the baseline.} The PID designs are not hand-tuned: their gains result from multistart searches on the same sampled models and under the same sensitivity, response and variance constraints as the LADRC designs, and they use the same measurements and the same translational feed-forward terms. Two outcomes of these searches should be kept in mind when the comparisons are read. First, the horizontal derivative filter is placed above the Nyquist frequency ($a_f\approx0$), so that no additional filtering is applied; the derivative action uses the measured velocity, not a differentiated position, and its noise contribution is part of the covariance model. Second, the common 10\% overshoot limit forces small set-point weights $b$ in several classes (Section~\ref{sec:req}), which penalizes the PID on moving references. The PID results therefore describe PID designs selected under the common specification. Baselines outside this specification, such as the default PX4 multicopter controllers, a PID with a reference prefilter or an INDI controller, were not evaluated.

\subsection{Actuator-aware, delay-aligned observer input (proposed)}

In the conventional commanded-input formulation, the observer is driven by
$\bar u_k=u_{a,k}$ for altitude and attitude in the nonlinear implementation, with $u_{a,k}=u_k$ in the unsaturated nominal linear model. Under \eqref{eq:chan}, however, the measured state at sample
$k$ corresponds to the plant state at $k-n_d$, whose acceleration is driven
by the earlier actuator output $a_{k-n_d}$ rather than by the current command
$u_k$. Consequently, on the delayed measurement time base, the effective
disturbance observed by the ESO is (in the nonlinear implementation with $u_k$ replaced by the limited input $u_{a,k}$)
\begin{equation}
	d_{k-n_d}+\underbrace{a_{k-n_d}-u_k}_{\text{observer-input mismatch}} .
	\label{eq:input_mismatch}
\end{equation}
The first term is the physical lumped disturbance; the second is a deterministic residual caused by actuator lag and measurement timing. The commanded-input formulation therefore forces the ESO to estimate both, so that rotor dynamics and measurement timing are absorbed into the
disturbance estimate. Increasing the observer bandwidth then forces the ESO to react more
rapidly to this input mismatch as well as to measurement noise, which can
increase control activity and erode closed-loop robustness.

For the altitude and attitude channels, the proposed actuator-aware
formulation instead propagates the limited virtual command through an internal
rotor-lag model,
\begin{equation}
	\hat a_{k+1}
	=
	\hat\alpha\,\hat a_k
	+(1-\hat\alpha)u_{a,k},
	\quad
	\bar u_k=\hat a_{k-\hat n_d},
	\quad
	\hat\alpha=e^{-T_s/\hat\tau_m},
	\label{eq:aa}
\end{equation}
where $u_{a,k}$ is the virtual input formed from the limited thrust or body torque before rotor allocation \eqref{eq:limited_input}, and a short history buffer of $\hat a_k$ provides the measurement-time alignment. The nominal implementations use $(\hat\tau_m,\hat n_d)=(30\,\mathrm{ms},5)$; A2-Unaligned uses $\hat n_d=0$. The alignment follows the delay-compensating observers reviewed in Section~\ref{sec:related}; at the nominal 5-ms delay its effect is small compared with that of the 30-ms rotor-lag model (Section~\ref{sec:abl}).

Under the nominal small-signal model \eqref{eq:chan} with matched $\hat\tau_m=\tau_m$ and $\hat n_d=n_d$, consistent initialization and no allocation-induced saturation, $\bar u_k$ is the modeled virtual acceleration acting at the time of the delayed measurement. The ESO then need not absorb the nominal rotor lag or measurement-time offset, and its error dynamics recover those of the delay-free ESO on the delayed time base. Componentwise command clipping does not break this reconstruction because \eqref{eq:aa} uses the limited input; rotor-lag or delay mismatch, individual-rotor saturation, nonlinear allocation and cross-channel coupling do. The physical lag and delay remain in the feedback path, so their effect on robustness is screened separately through $M_s$.

For the horizontal channels, the observer is driven by the near-hover
horizontal-acceleration surrogate reconstructed from the measured attitude,
\begin{equation}
	\begin{aligned}
		\bar u_x &=
		g(\theta_m\cos\psi_m+\phi_m\sin\psi_m),\\
		\bar u_y &=
		g(\theta_m\sin\psi_m-\phi_m\cos\psi_m).
	\end{aligned}
	\label{eq:att}
\end{equation}
Because $(\phi_m,\theta_m,\psi_m)$ are delayed measurements of the realized
attitude, these inputs implicitly contain the closed inner-loop and rotor
dynamics and are aligned with the same measurement time base as the horizontal
position observations. This measured-attitude construction avoids replacing
the realized horizontal acceleration by the instantaneous outer-loop virtual
command. The inputs \eqref{eq:att} are near-hover acceleration surrogates reconstructed from the delayed attitude, not acceleration measurements; their accuracy degrades away from the small-angle regime of \eqref{eq:inv2}.

\subsection{Discrete-time channel model}
For each controlled channel, the sampled channel model in
\eqref{eq:chan} is combined with the implemented controller and observer to
form a discrete linear state-space representation. For the horizontal
channels $x$ and $y$, this representation additionally includes the complete
closed attitude loop, including its controller, actuator lag, and measurement
delay. The resulting model is written as
\begin{equation}
	\begin{aligned}
		\mathbf{X}_{k+1}
		&=
		\boldsymbol{\Phi}\mathbf{X}_k
		+\boldsymbol{\Gamma}_d d_k
		+\boldsymbol{\Gamma}_n\mathbf{n}_k,\\
		e_k
		&=
		\mathbf{c}_e\mathbf{X}_k,\qquad
		u_k=
		\mathbf{c}_u\mathbf{X}_k+\mathbf{d}_u\mathbf{n}_k ,
	\end{aligned}
	\label{eq:dss}
\end{equation}
where $\mathbf{X}_k\in\mathbb R^{n_x}$ is the perturbation state about a fixed reference and nominal hover (plant, actuator, observer, controller and delay states, excluding $d_k$), $e_k=\chi_k-\chi_{d,k}$, and $\mathbf c_e,\mathbf c_u,\mathbf d_u$ are row vectors. The matrices are extracted column by column from the same discrete update functions used in the nonlinear simulation, so the covariance analysis reflects the implemented controller.

The lumped disturbance is represented as a discrete random walk,
\begin{equation}
	d_{k+1}=d_k+w_k,
	\qquad
	\mathbb{E}\{w_k^2\}=q_dT_s ,
	\label{eq:dist_rw}
\end{equation}
with $w_k$ zero-mean white and independent of the white measurement noise $\mathbf n_k$; $d_k$ enters through $\boldsymbol{\Gamma}_d$ (position and rate components $T_s^2/2$ and $T_s$); in the horizontal cascade model it enters the translational channel only. Translational states use a zero-order-hold acceleration, whereas the actuator retains the exact pole $\alpha=e^{-T_s/\tau_m}$, so \eqref{eq:dss} is a sampled-data approximation tailored to the implemented controller. It excludes drag, nonlinear cross-axis coupling, saturation and gyro-bias dynamics; the predictions below are tuning constraints, not stochastic guarantees for the nonlinear plant.

\subsection{Variances and noise-induced control-activity budget}
Let
\[
\boldsymbol{\Sigma}_n
=
\operatorname{diag}
\!\left(\sigma_{n,1}^2,\ldots,\sigma_{n,n_n}^2\right)
\]
denote the covariance of the independent measurement-noise inputs in the
nominal hover model; $\mathbf I_{n_x}$ denotes the identity matrix, distinct from the physical inertia matrix $\mathbf I$. For a Schur-stable $\boldsymbol{\Phi}$, the
noise-induced steady-state covariance $\mathbf{P}_n$ is the unique solution of
the discrete Lyapunov equation
\begin{equation}
	\mathbf{P}_n
	=
	\boldsymbol{\Phi}\mathbf{P}_n\boldsymbol{\Phi}^{\top}
	+
	\boldsymbol{\Gamma}_n
	\boldsymbol{\Sigma}_n
	\boldsymbol{\Gamma}_n^{\top}.
	\label{eq:covlyap}
\end{equation}
The corresponding tracking-error and control variances are
\begin{equation}
	\begin{aligned}
		\sigma_{e,n}^2
		&=
		\mathbf{c}_e\mathbf{P}_n\mathbf{c}_e^{\top},\\
		\sigma_u^2
		&=
		\mathbf{c}_u\mathbf{P}_n\mathbf{c}_u^{\top}
		+
		\mathbf{d}_u\boldsymbol{\Sigma}_n\mathbf{d}_u^{\top},
	\end{aligned}
	\label{eq:lyap}
\end{equation}
where the second term in $\sigma_u^2$ accounts for direct measurement-noise
feedthrough to the implemented control law.

To constrain rapid actuator-command variation, a further design metric is the variance of the sample-to-sample increment $\Delta u_k=u_k-u_{k-1}$, whose differencing weight $4\sin^2(\omega T_s/2)$ emphasizes high-frequency noise-induced activity. Using
$\mathbf{X}_k=
\boldsymbol{\Phi}\mathbf{X}_{k-1}
+\boldsymbol{\Gamma}_n\mathbf{n}_{k-1}$
for the noise-driven model gives
\begin{equation}
	\begin{aligned}
		\sigma_{\Delta u}^2
		&=
		\mathbf{h}_{\Delta}\mathbf{P}_n\mathbf{h}_{\Delta}^{\top}
		+
		\mathbf{r}_{\Delta}\boldsymbol{\Sigma}_n
		\mathbf{r}_{\Delta}^{\top}
		+
		\mathbf{d}_u\boldsymbol{\Sigma}_n\mathbf{d}_u^{\top},\\
		\mathbf{h}_{\Delta}
		&=
		\mathbf{c}_u(\boldsymbol{\Phi}-\mathbf{I}_{n_x}),\\
		\mathbf{r}_{\Delta}
		&=
		\mathbf{c}_u\boldsymbol{\Gamma}_n-\mathbf{d}_u ,
	\end{aligned}
	\label{eq:du}
\end{equation}
where $\mathbf{X}_{k-1}$, $\mathbf{n}_{k-1}$, and $\mathbf{n}_k$ are mutually
uncorrelated under the white-noise model.

Under the random-walk model \eqref{eq:dist_rw}, because $d_k$ itself is nonstationary, Schur stability of
$\boldsymbol{\Phi}$ alone does not give a finite tracking-error variance: the closed-loop dc gain from $d$ to $e$
must also vanish, so that the disturbance level does not appear in the tracking error,
\begin{equation}
	\mathbf{c}_e
	(\mathbf{I}_{n_x}-\boldsymbol{\Phi})^{-1}
	\boldsymbol{\Gamma}_d=0 .
	\label{eq:dc_rejection}
\end{equation}
This condition is checked numerically for every stable candidate retained in
the tuning procedure. Defining
\begin{equation}
	\bar{\mathbf{x}}
	=
	(\mathbf{I}_{n_x}-\boldsymbol{\Phi})^{-1}\boldsymbol{\Gamma}_d ,
\end{equation}
the disturbance-induced tracking variance is
\begin{equation}
	\sigma_{e,d}^2
	=
	q_dT_s\,
	\bar{\mathbf{x}}^{\top}\mathbf{Q}\bar{\mathbf{x}},
	\qquad
	\mathbf{Q}
	=
	\boldsymbol{\Phi}^{\top}\mathbf{Q}\boldsymbol{\Phi}
	+
	\mathbf{c}_e^{\top}\mathbf{c}_e .
	\label{eq:dvar}
\end{equation}
It follows from the stable transformed state $\mathbf X^{\circ}_k=\mathbf X_k-\bar{\mathbf x}d_k$, which obeys $\mathbf X^{\circ}_{k+1}=\boldsymbol\Phi\mathbf X^{\circ}_k-\bar{\mathbf x}w_k$ with $e_k=\mathbf c_e\mathbf X^{\circ}_k$ by \eqref{eq:dc_rejection}; it measures the accumulated transient response to successive random-walk increments. Designs that violate \eqref{eq:dc_rejection} are not assigned a stationary disturbance variance.

\subsection{Sensitivity and response requirements}
\label{sec:req}

Closed-loop sensitivity is evaluated only for Schur-stable candidates by
injecting a perturbation $\delta u_k$ at the virtual-input summing point and taking the total plant input $u_k+\delta u_k$ as output. With all other exogenous inputs zero, define $S_{\rm in}(z)=[U(z)+\Delta U(z)]/\Delta U(z)$ and $M_s=\sup_{0\le\omega\le\pi/T_s}|S_{\rm in}(e^{\mathrm j\omega T_s})|$, where $\mathrm j^2=-1$; equivalently $S_{\rm in}=1/(1+\Lin)$ with $\Lin$ the loop gain broken at that point. $M_s$ is a per-loop single-input single-output (SISO) quantity: for $z$ and the attitude classes the loop is broken at the channel virtual input, and for $xy$ at the outer virtual input $u_x$ with the designed attitude loop closed inside the cascade model. It is a sampled-loop sensitivity screen, not a multi-input multi-output (MIMO) singular-value margin, and it is not a proof of coupled MIMO or nonlinear robust stability. During the
design search, $M_s$ is evaluated on a 600-point logarithmic frequency grid with local peak refinement, and final candidates on 4000 points (Appendix~\ref{app:search}). The tuning requirement is $M_s\leq1.98$; final candidates are audited against the reporting threshold $M_s\leq2$. Sensitivity peaks of at most about 2 are a common robustness guideline in PID design~\cite{astrom2006}.

The response requirement is evaluated on the nominal sampled closed-loop linear model of each class, for $xy$ on the cascade model with the designed attitude loop closed inside. For the unit reference step over a 400-s horizon, the overshoot $\mathrm{OS}=100\max\{0,\max_k\chi_k-1\}$ must not exceed 10\% and the 2\% settling time $t_s$ must not exceed $\bar t_{s,c}=15$, 5, 2 and 5~s for $c=xy,z,\phi\theta,\psi$. The settling bound prevents a loop that is too slow from passing the overshoot screen; both bounds were fixed before the redesigns reported in Section~\ref{sec:results}. These linear screens do not bound arbitrary nonlinear transients. For the PID designs, the set-point weight $b$ is set per class to the largest value of the grid $\{0,0.05,\ldots,1\}$, not below the searched value, that meets the overshoot and settling requirements. For the $xy$, $z$ and $\psi$ classes, whose references are exogenous, $b$ acts only on the reference response. In the unsaturated linear model (where the clamp of \eqref{eq:pidstate} is inactive), it leaves the loop, $M_s$ and the noise covariances unchanged. The attitude weights are different: their reference is the noisy tilt command, so they would change the cascade covariance, and they are retained. The rule leaves them at their searched values in all three PID designs, and the recomputed $M_s$, variances and increments are unchanged (\texttt{setpoint\_invariance.csv}). For PID-Budget this raises $b$ from 0.19 to 0.35 ($xy$) and from 0.04 to 0.05 ($\psi$). For PID-Budget-S and PID-E the overshoot limit is already active at the searched $b$ for the selected PID gains, so $b$ remains of order $10^{-3}$ in several classes, and their tracking of moving references is limited by the response specification that is common to all designs.

For LADRC, in each symmetric channel class
$c\in\{xy,z,\phi\theta,\psi\}$, the controller and observer bandwidths are
selected to minimize the predicted steady-state tracking variance,
\begin{equation}
	\begin{aligned}
		\min_{\omega_c,\,\omega_o}\quad
		& \sigma_e^2
		=\sigma_{e,d}^2+\sigma_{e,n}^2 \\
		\text{s.t.}\quad
		& \sigma_u\leq\bar u_c, \\
		& \sigma_{\Delta u}\leq\overline{\Delta u}_c, \\
		& M_s\leq1.98, \\
		& \mathrm{OS}\leq10\%,\quad t_s\leq\bar t_{s,c}, \\
		& \omega_oT_s\leq0.2, \\
		& \sigma_{\Delta u_i,\mathrm{cas}}
		\leq B_{\mathrm{cas}},
		\qquad c=xy\ \text{when reselected}.
	\end{aligned}
	\label{eq:opt}
\end{equation}
Here, $\sigma_u$ and $\sigma_{\Delta u}$ are the noise-induced standard
deviations predicted by \eqref{eq:lyap} and \eqref{eq:du}, respectively,
whereas $\sigma_{\Delta u_i,\mathrm{cas}}$ constrains the corresponding
inner-loop control increment generated by noise propagated through the
complete horizontal cascade. The local quantities $\sigma_u$ and $\sigma_{\Delta u}$ and their ceilings are always evaluated with the white-noise model of Table~\ref{tab:params}. In contrast, $\sigma_e^2$ and $\sigma_{\Delta u_i,\mathrm{cas}}$ use the noise model of the loop being designed: white for the MATLAB designs and estimator-coloured for the -E designs of Section~\ref{sec:estaware}. $M_s$, OS and $t_s$ do not depend on the noise model.

The inner attitude and altitude classes are designed first; their closed-loop models are then embedded in the horizontal covariance model, so that the $xy$ design includes the realized inner-loop dynamics. The search over $r=\omega_o/\omega_c$ and $\omega_c$ is summarized in Appendix~\ref{app:search}.

The control-standard-deviation ceilings $\bar u_c$ are defined in normalized
virtual-input coordinates. For the horizontal channels, the ceiling
corresponds to a $0.5^{\circ}$ tilt-command equivalent. The altitude ceiling
is $2\%$ of the available thrust margin; the roll/pitch ceiling is $2\%$ of
a $0.41$-N\,m differential-torque reference, $lk(\omega_{\max}^2-\omega_h^2)=0.412$~N\,m, obtained with one rotor of the pair at $\omega_{\max}$ and the other at the hover speed $\omega_h=\sqrt{mg/(4k)}$; and the yaw ceiling is $2\%$
of the yaw-torque limit. These normalized tuning limits differ from the physical 0.2-N\,m roll/pitch torque limit: the $0.5^{\circ}$ and $2\%$ levels are engineering-selected tuning budgets, not actuator limits. The local increment ceilings below are PID-referenced, and the random-walk intensities $q_{d,c}$ are engineering-selected weights of the variance objective, not disturbance statistics identified from the evaluation scenarios.

For each LADRC candidate, the local increment ceiling
$\overline{\Delta u}_c$ is referenced to the corresponding noise-induced
increment obtained with PID-Budget. PID-Budget itself is the reference and is not constrained by it; A3 omits the local ceiling.
Let $s_{d,c}$ denote the random-walk standard-deviation intensity for channel
class $c$, so that $q_{d,c}=s_{d,c}^2$ in \eqref{eq:dist_rw} and
\eqref{eq:dvar}. The values used for
$c\in\{xy,z,\phi\theta,\psi\}$ are
\[
\left[
\frac{0.3}{m},\;
\frac{0.3}{m},\;
\frac{0.01}{I_{xx}},\;
\frac{0.002}{I_{zz}}
\right],
\]
in force or torque per $\sqrt{\mathrm{s}}$ divided by mass or inertia.

For the horizontal design, the complete cascade model additionally exposes
the inner virtual control as
\begin{equation}
	u_{i,k}
	=
	\mathbf{c}_i\mathbf{X}_k
	+\mathbf{d}_i\mathbf{n}_k .
	\label{eq:inner_output}
\end{equation}
Replacing $(\mathbf{c}_u,\mathbf{d}_u)$ in \eqref{eq:du} with
$(\mathbf{c}_i,\mathbf{d}_i)$ and using the full measurement-noise covariance
$\boldsymbol{\Sigma}_n$, including position, velocity, attitude, and rate
noise, yields the cascade-level increment prediction
$\sigma_{\Delta u_i,\mathrm{cas}}$. This quantity captures the path by which
outer-loop measurement noise propagates through commanded tilt into
inner-loop control activity.
It is imposed during horizontal reselection of the AA and ablation designs, all strict candidates and every budget-curve candidate, as a stationary nominal-hover tuning constraint rather than a bound on nonlinear rotor-command increments.

At zero-yaw hover, the symmetry of the model and gain selection gives
identical one-axis covariance predictions for $x/y$ and for roll/pitch.
Accordingly, one representative channel is used for each symmetric class.

Two cascade thresholds are examined (Section~\ref{sec:cascade}): the \emph{PID-referenced} threshold $B_{\mathrm{cas}}=1.382$ is the inner-loop noise-only hover increment of PID-Budget, and a second, deliberately stricter common threshold, $B_{\mathrm{cas}}=0.598$ (\emph{strict}), is defined from the isolated-attitude PID increment. It serves to compare the controllers under a common, tighter activity requirement, not to identify an optimal threshold. The reported parameters are the best feasible designs found by the stated finite search (Appendix~\ref{app:search}); no claim of global optimality is made.

\subsection{Estimator-aware noise model}\label{sec:estaware}
The ceilings above assume white noise at the controller input. When the controller is fed by a state estimator, $\mathbf n_k$ in \eqref{eq:dss} is the estimation error, which is coloured and correlated across the measured signals. It is represented by a stationary shaping filter driven by unit white noise $\mathbf v_k$,
\begin{equation}
	\mathbf z_{k+1}=\mathbf A_f\mathbf z_k+\mathbf B_f\mathbf v_k,\qquad
	\mathbf n_k=\mathbf C_f\mathbf z_k+\mathbf D_f\mathbf v_k,
	\label{eq:shaping}
\end{equation}
which is appended to the channel state, $\widetilde{\mathbf X}_k=[\mathbf X_k^{\top}\ \mathbf z_k^{\top}]^{\top}$:
\begin{equation}
	\begin{aligned}
		\widetilde{\boldsymbol\Phi}&=\begin{bmatrix}\boldsymbol\Phi&\boldsymbol\Gamma_n\mathbf C_f\\ \mathbf 0&\mathbf A_f\end{bmatrix},\qquad
		\widetilde{\boldsymbol\Gamma}_n=\begin{bmatrix}\boldsymbol\Gamma_n\mathbf D_f\\ \mathbf B_f\end{bmatrix},\\
		\widetilde{\mathbf c}_u&=\begin{bmatrix}\mathbf c_u&\mathbf d_u\mathbf C_f\end{bmatrix},\qquad
		\widetilde{\mathbf d}_u=\mathbf d_u\mathbf D_f,
	\end{aligned}
	\label{eq:aug}
\end{equation}
and analogously for $\mathbf c_e$ and $(\mathbf c_i,\mathbf d_i)$. Equations \eqref{eq:covlyap}--\eqref{eq:du} and \eqref{eq:inner_output} then apply unchanged with $\boldsymbol\Sigma_n=\mathbf I$, because $\widetilde{\mathbf X}_{k-1}$, $\mathbf v_{k-1}$ and $\mathbf v_k$ are again uncorrelated. With $\mathbf A_f$ empty and $\mathbf D_f=\boldsymbol\Sigma_n^{1/2}$ the white-noise model is recovered (relative difference below $10^{-14}$ for all reported designs). A measured signal that lags the true one, e.g.\ after a rate filter, keeps its own delay buffer in \eqref{eq:dss}. The filter must be identified for the estimator and sensor configuration at hand; Section~\ref{sec:sitl} identifies it from PX4 SITL logs. Because the estimation error also depends on the vehicle motion and hence on the controller, a filter identified with one set of designs is an approximation for another, and a new design can be followed by a refit. All identified filters are stationary (spectral radius of $\mathbf A_f$ below one).

\emph{Delay convention.} On the PX4 stack the three body rates are filtered and lag the plant by about 6~ms, whereas the PX4 extended Kalman filter (EKF2)~\cite{px4ekf2} position, velocity and attitude estimates show no material lag (Section~\ref{sec:sitl}). The models of the -E designs therefore keep a 6-ms buffer on all body rates and none on the other signals; this convention is used for $M_s$, the step response and the covariances.

\emph{Rotor-command increment.} With the hover-linearized ``+'' mixer, $\Delta\boldsymbol\omega_{\mathrm{cmd},k}=\mathbf M\Delta\mathbf u_{v,k}$ with $\mathbf u_v=[u_z,u_\phi,u_\theta,u_\psi]^\top$, so $\operatorname{Cov}(\Delta\boldsymbol\omega_{\mathrm{cmd}})=\mathbf M\operatorname{Cov}(\Delta\mathbf u_v)\mathbf M^\top$ and the predicted increment of \eqref{eq:reported_metrics} is $J^2=\operatorname{tr}\{\mathbf M\operatorname{Cov}(\Delta\mathbf u_v)\mathbf M^\top\}/4$. The columns $\mathbf m_j$ of $\mathbf M$ are mutually orthogonal for the ``+'' configuration, so $\mathbf M^\top\mathbf M$ is diagonal and $J^2=\tfrac14\sum_{j}\|\mathbf m_j\|^2\sigma_{\Delta u_j}^2$, $j\in\{z,\phi,\theta,\psi\}$: cross-covariances between the classes do not change $J$ (checked numerically on the logged SITL increments, where the linear formula reproduces $J$ within 0.06\%).

\emph{Estimator-aware redesign.} A design for an estimator-fed loop (suffix -E) follows four steps. (i) The ceilings are fixed before any redesign as the estimator-aware roll/pitch-cascade and yaw increments of PID-Budget, $B^{E}_{\mathrm{cas}}=0.0065$ and $B^{E}_{\psi}=4.75\times10^{-4}$, computed with the model pooled over the designs already logged. (ii) Each class is screened as in \eqref{eq:opt}, with the delay convention above, $M_s\le1.98$, the response requirement, the white-noise local ceilings, $\sigma_{\Delta u_i,\mathrm{cas}}\le0.99B^{E}_{\mathrm{cas}}$ and, for yaw, $\sigma_{\Delta u_\psi}\le0.99B^{E}_{\psi}$; the altitude class has no estimator-aware ceiling, and a class of the nominal design that already meets these constraints is kept. (iii) The new designs are logged in noise-only hover, the pooled model is refitted with their logs, and the constraints are rechecked. (iv) If a constraint is violated, steps (ii)--(iii) are repeated, at most twice; otherwise the gains are frozen. Step (iii) tests whether the noise model used in step (ii) remains valid for the closed loops it produced.

\subsection{Nominal candidates and implementation}
Four nominal controller candidates are evaluated using the same nonlinear
command inversion, rotor mixer, actuator dynamics, and sensor model.
\emph{PID-Budget} is a two-degree-of-freedom cascade PID baseline.
\emph{AA-Budget} uses the proposed actuator-aware LADRC formulation with
regular reduced-order ESOs. \emph{RO-HalfGain} replaces the regular observers
with predictive half-gain observers while retaining commanded inputs for the
altitude and attitude observers. \emph{AA-HalfGain} combines the same
half-gain observer structure with the proposed actuator-aware inputs. Both
half-gain variants retain the measured-attitude input formulation for the
horizontal observers. Because each candidate is independently tuned subject
to the applicable activity constraints and sensitivity screens, their nominal
comparison reflects complete controller designs rather than an input-only,
fixed-gain ablation.

The PID-Budget gains are obtained by multistart optimization of the inner loops, followed by outer-loop and joint-cascade searches. The PID law is \eqref{eq:pid}--\eqref{eq:pidstate} (Section~\ref{sec:pid}). Table~\ref{tab:protocol} summarizes the design protocol, and Table~\ref{tab:gains} reports the implemented gains.

\subsection{Predictive half-gain observer adaptation}
For the rate--disturbance model
\[
\mathbf{A}_d=
\begin{bmatrix}
	1&T_s\\
	0&1
\end{bmatrix},
\]
the discrete-time half-gain pole placement of \cite{herbst2025halfgain} is
adapted here to the predictive observer timing used in the implementation.
Let
\[
h=\frac{\omega_oT_s}{2},
\qquad
\lambda_o=e^{-h},
\qquad
\varepsilon_{v,k}=\dot\chi_{m,k}-\hat v_k .
\]
The observer update is
\begin{equation}
	\begin{aligned}
		\hat v_{k+1}
		&=
		\hat v_k+T_s(\hat d_k+\bar u_k)+L_1\varepsilon_{v,k},\\
		\hat d_{k+1}
		&=
		\hat d_k+L_2\varepsilon_{v,k},\\
		L_1
		&=
		2-2\lambda_o\cos h,\\
		L_2
		&=
		\frac{1-2\lambda_o\cos h+\lambda_o^2}{T_s}.
	\end{aligned}
	\label{eq:hg_predictive}
\end{equation}
These gains place the discrete observer-error poles at $p_{1,2}=\exp[(-1\pm \mathrm{j})\omega_oT_s/2]$; with the current-observer gain $\mathbf{L}_c$, the predictive form is $\mathbf{L}=\mathbf{A}_d\mathbf{L}_c$. As $T_s\rightarrow0$, $L_1/T_s\rightarrow\omega_o$ and $L_2/T_s\rightarrow\omega_o^2/2$, the reduced gains characteristic of the half-gain construction, whereas the regular RESO \eqref{eq:reso} keeps the forward-Euler bandwidth gains.

Equation~\eqref{eq:hg_predictive} is a rate-aided predictive adaptation of
the half-gain observer principle in \cite{herbst2025halfgain}, rather than a
replication of the complete controller proposed therein. \emph{RO-HalfGain}
uses the commanded virtual input for the altitude and attitude observers,
whereas \emph{AA-HalfGain} uses the delay-aligned actuator-aware input in
\eqref{eq:aa}. Both variants use the measured-attitude formulation
\eqref{eq:att} for the horizontal observers and are tuned subject to the
same local variance, control-increment, sensitivity, and overshoot
constraints. AA-HalfGain additionally enforces the cascade-level inner-control increment constraint during horizontal bandwidth selection.

\section{Results}\label{sec:results}
For an evaluation window containing $N$ consecutive samples, the per-run metrics are
\begin{align}
\mathrm{RMSE}_{3D}&=\sqrt{\frac1N\sum_{k=1}^{N}\|\boldsymbol\xi_{d,k}-\boldsymbol\xi_k\|_2^2},\nonumber\\
J_{\Delta\wcmd}&=\sqrt{\frac{1}{4(N-1)}\sum_{i=1}^{4}\sum_{k=2}^{N}(\wcmd[,i,k]-\wcmd[,i,k-1])^2}.
\label{eq:reported_metrics}
\end{align}
The reported rotor-command increment is the mean of the per-run $J_{\Delta\wcmd}$ values; it is a root-mean-square (RMS) increment in rad/s, not divided by $T_s$, and differs from the noise-only virtual-control increment used in tuning. Across $N_r$ bounded runs, a mean RMSE confidence interval has half-width $t_{0.975,N_r-1}s_R/\sqrt{N_r}$, where $s_R$ is the sample standard deviation; paired comparisons apply the same formula to within-seed differences of jointly bounded pairs.
All nonlinear trials use the same 16-state rigid-body/rotor model, fourth-order Runge--Kutta (RK4) integration and controller/sensor period $T_s=1$ ms; scenarios S5 and S6 are defined in Appendix~\ref{app:scenarios}. Nominal and ablation tests use thirty paired seeds 91001--91030, and budget curves the same thirty seeds at each level. Noise/latency tests use ten pairs 93001--93010 per condition; mismatch tests use ten pairs 92001--92010 per plant. At each seed, controllers share sensor samples, gyro bias and turbulence. A failure is a nonfinite state or any position component exceeding 50 m; statistics use bounded runs only. A rotor-fluctuation threshold exceedance is $\sqrt{\mathrm{mean}_i\operatorname{var}(\omega_i)}>20$ rad/s in a bounded run; failures and exceedances are disjoint counts. Bold entries mark results discussed in the text and do not imply statistical significance or an overall ranking.

\subsection{Controller protocol and implemented gains}
Tables~\ref{tab:protocol} and \ref{tab:gains} specify the controller families and the gains used in all nominal nonlinear comparisons. PID-Budget retains its nominal gains except the set-point weights (Section~\ref{sec:req}); AA-Budget, AA-HalfGain, A2 and A3 undergo horizontal reselection with the cascade increment ceiling, as do the strict designs and every budget-curve candidate. The response requirement of Section~\ref{sec:req} is imposed on every design. For PID-Budget-S, RO-HalfGain, A1-Commanded and the PID budget-curve points $\gamma=0.5$ and 0.75, meeting it required the dedicated searches of Appendix~\ref{app:search}. RO-HalfGain and A1-Commanded are searched for every class on a grid of observer/controller ratios $1\le\omega_o/\omega_c\le60$, with the primary cascade ceiling for $xy$. The nominal AA input models use $\hat\tau_m=30$ ms and $\hat n_d=5$; full-precision gains, including strict and budget-curve designs, accompany the code.

\begin{table*}[t]\centering\scriptsize
\caption{Controller Design Protocol}\label{tab:protocol}
\setlength{\tabcolsep}{4pt}\begin{tabular}{lllll}\toprule
Controller & Architecture & Common constraints & Increment ceiling & Tuning\\\midrule
PID-Budget & two-degree-of-freedom PID & variance, $M_s$, OS, $t_s$ & none & multistart + joint cascade\\
AA-Budget & actuator-aware RO-LADRC & variance, $M_s$, OS, $t_s$ & PID local value & bandwidth/ratio selection\\
RO-HalfGain & commanded inner/altitude ESO & variance, $M_s$, OS, $t_s$ & PID local value & predictive half-gain + bandwidth\\
AA-HalfGain & actuator-aware half-gain ESO & variance, $M_s$, OS, $t_s$ & PID local value & predictive half-gain + bandwidth\\
A1-Commanded & commanded inner/altitude RO-ESO & same as AA-Budget & same & redesign\\
A2-Unaligned & lag-model RO-ESO, $\hat n_d=0$ & same as AA-Budget & same & redesign\\
A3-NoIncrement & actuator-aware RO-ESO & same as AA-Budget & removed & redesign\\
A4-CommandedTilt & AA inner/altitude, commanded-tilt $x,y$ ESO & same as AA-Budget & same & fixed / horizontal redesign\\
\bottomrule\end{tabular}\par\smallskip\begin{minipage}{\linewidth}\footnotesize The full-cascade increment limit applies to the horizontal reselection of AA-Budget, AA-HalfGain and A1--A3 and to the strict and budget-curve searches; RO-HalfGain and A1 are searched in all classes. AA internal model: 30-ms lag, 5-ms alignment (A2: none). $M_s$ target 1.98 (reporting 2), OS 10\%, settling as in Section~\ref{sec:req}.\end{minipage}
\end{table*}

\begin{table*}[!t]
	\centering
	\scriptsize
	\caption{Implemented Controller Gains}
	\label{tab:gains}
	\setlength{\tabcolsep}{3pt}
	\renewcommand{\arraystretch}{1.08}
	
	\begin{minipage}[t]{0.33\textwidth}
		\centering
		\textbf{(a) PID-Budget}\\[2pt]
		\setlength{\tabcolsep}{2pt}
		\begin{tabular}{lrrrrr}
			\toprule
			Class & $K_p$ & $K_i$ & $K_d$ & $\omega_f$ & $b$\\
			\midrule
			$xy$          & 4.465   & 3.358   & 3.215  & 15362.828 & 0.350\\
			$z$           & 19.460  & 32.427  & 6.990  & 21.076    & 0.353\\
			$\phi\theta$  & 120.609 & 145.653 & 19.056 & 449.779   & 0.914\\
			$\psi$        & 32.212  & 68.577  & 9.880  & 34.865    & 0.050\\
			\bottomrule
		\end{tabular}
	\end{minipage}
	\hfill
	\begin{minipage}[t]{0.65\textwidth}
		\centering
		\textbf{(b) LADRC bandwidths $(\omega_c,\omega_o)$ [rad/s]}\\[2pt]
		\begin{tabular}{lrrrr}
			\toprule
			Controller
			& \multicolumn{1}{c}{$xy$}
			& \multicolumn{1}{c}{$z$}
			& \multicolumn{1}{c}{$\phi\theta$}
			& \multicolumn{1}{c}{$\psi$}\\
			\midrule
			AA-Budget
			& $(2.057,31.700)$
			& $(3.786,43.466)$
			& $(10.426,195.136)$
			& $(5.031,74.498)$\\
			
			RO-HalfGain
			& $(0.786,38.859)$
			& $(1.258,13.155)$
			& $(3.905,3.905)$
			& $(1.258,13.155)$\\
			
			AA-HalfGain
			& $(2.154,40.318)$
			& $(3.816,56.503)$
			& $(10.908,183.422)$
			& $(5.106,95.563)$\\
			
			A1-Commanded
			& $(0.760,30.939)$
			& $(2.120,18.252)$
			& $(3.905,3.905)$
			& $(2.120,18.252)$\\
			
			A2-Unaligned
			& $(2.056,31.698)$
			& $(3.882,40.583)$
			& $(10.426,195.150)$
			& $(4.969,76.597)$\\
			
			A3-NoIncrement
			& $(2.194,27.854)$
			& $(3.889,40.656)$
			& $(12.372,190.707)$
			& $(4.974,76.672)$\\
			\bottomrule
		\end{tabular}
	\end{minipage}\par\smallskip\begin{minipage}{\linewidth}\footnotesize (a) PID $[K_p,K_i,K_d,\omega_f,b]$ of \eqref{eq:pid}--\eqref{eq:pidstate}, $\omega_f$ in rad/s; (b) LADRC $(\omega_c,\omega_o)$ in rad/s. $x/y$ and roll/pitch share gains; the horizontal $\omega_f$ gives $a_f\simeq2.13\times10^{-7}$. Strict and -E gains: Table~\ref{tab:gains2}.\end{minipage}
	
\end{table*}

\subsection{Actuator-input mechanism and nominal performance}\label{sec:nominal}
\begin{figure*}[tbp]\centering\includegraphics[width=.84\textwidth,height=.18\textheight,keepaspectratio]{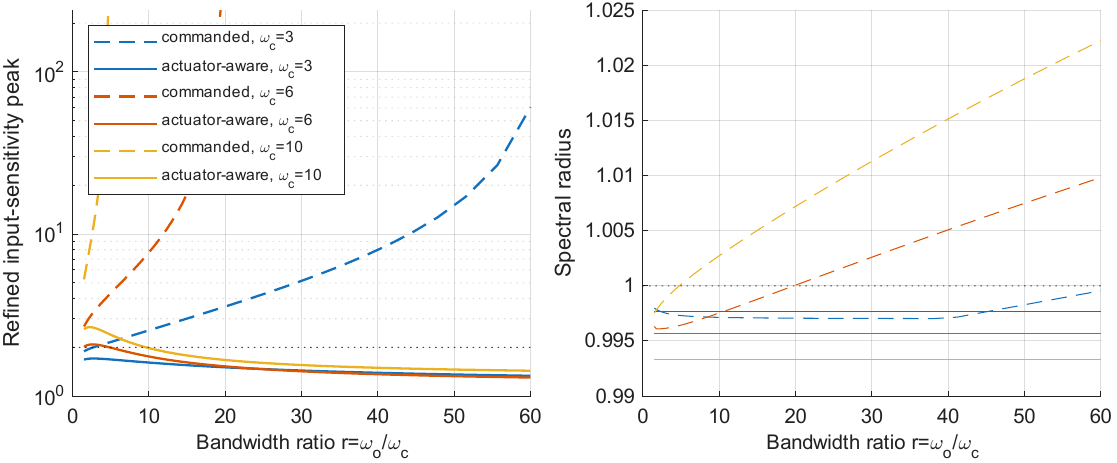}\caption{Roll-channel sweep of the observer/controller bandwidth ratio $\omega_o/\omega_c$ at $\omega_c=3$, 6 and 10~rad/s with commanded (dashed) and actuator-aware (solid) observer inputs. Left: input-sensitivity peak $M_s$ (dotted line: $M_s=2$); gaps mark loops that are not Schur stable. Right: spectral radius (dotted line: 1).}\label{fig:msratio}\end{figure*}

\begin{figure*}[tbp]\centering\includegraphics[width=.84\textwidth,height=.2\textheight,keepaspectratio]{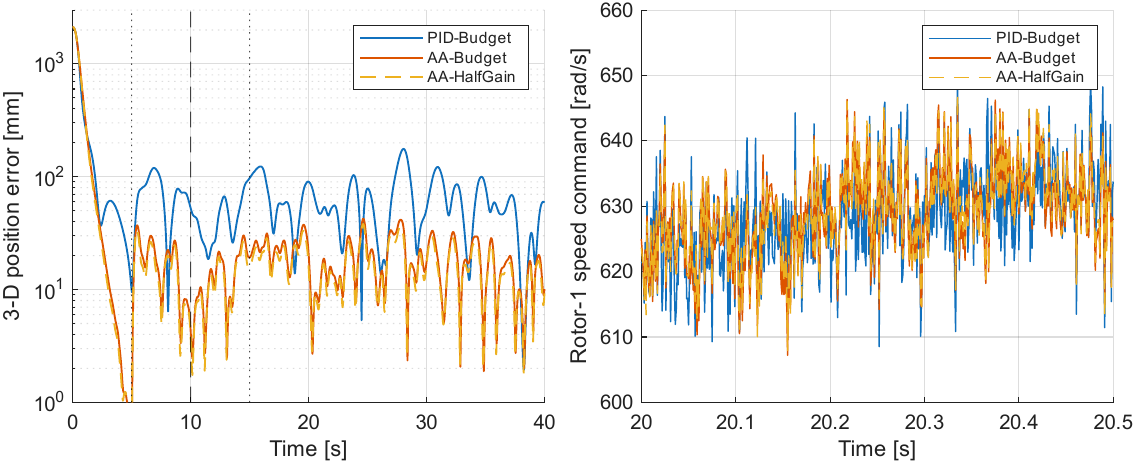}\caption{Time-domain responses in one paired S6 trial (seed 91001). Left: norm of the 3-D position error on a logarithmic scale; the dotted lines mark the onset of the wind (5~s) and of the torque disturbance (15~s), and the dashed line the start of the metric window (10~s). Right: rotor-1 speed command over 0.5~s. In this trial the RMSEs of PID-Budget, AA-Budget and AA-HalfGain are 67.9, 20.1 and 17.7~mm and their rotor-command increments 9.93, 7.47 and 7.59~rad/s.}\label{fig:timedomain}\end{figure*}

\begin{table}[t]\centering\scriptsize
\caption{Nominal Set-Point Performance (S6, 30 Paired Trials)}\label{tab:nominal}
\setlength{\tabcolsep}{2.2pt}\begin{tabular}{lrrrrrr}\toprule
Controller & $\sigma_{xy}$ & $M_s$ & RMSE & CI & Inc. & Fail\\
 & [mm] & & mean$\pm$SD [mm] & [mm] & [rad/s] & \\\midrule
PID-Budget & 132.9 & 1.978 & 65.5$\pm$7.0 & 2.62 & 9.91 & 0/30\\
AA-Budget & 32.2 & 1.974 & 19.2$\pm$1.5 & 0.58 & 7.46 & 0/30\\
RO-HalfGain & 300.5 & 1.963 & 226.4$\pm$13.4 & 5.00 & 0.80 & 0/30\\
AA-HalfGain & \keyresult{27.8} & 1.974 & \keyresult{16.9$\pm$1.3} & 0.50 & 7.58 & 0/30\\
\bottomrule\end{tabular}\par\smallskip\begin{minipage}{\linewidth}\footnotesize $\sigma_{xy}$: predicted horizontal tracking SD; RMSE: nonlinear 3-D position error, mean$\pm$SD; CI: 95\% half-width; $M_s$: maximum over the channel classes.\end{minipage}\end{table}

Fig.~\ref{fig:msratio} isolates the observer input at fixed bandwidths $\omega_c=3$, 6 and 10 rad/s, sweeping $\omega_o/\omega_c$ with commanded (dashed) and actuator-aware (solid) inputs; gaps in the $M_s$ curves mark non-Schur loops, as the spectral radius shows. With commanded inputs the roll-channel sensitivity peak grows with the observer/controller ratio and, for $\omega_c=6$ and 10 rad/s, the loop loses Schur stability, whereas with actuator-aware inputs the peak decreases with the ratio and the loop remains Schur stable. Table~\ref{tab:nominal} compares the complete, independently tuned designs, so its differences combine the observer input, the feasible bandwidths and the activity constraints; the effect of the observer input itself is isolated by Fig.~\ref{fig:msratio} and by the ablations of Section~\ref{sec:abl}. AA-Budget reduces mean position RMSE by 70.7\% relative to PID-Budget while reducing measured rotor-command increments. Fig.~\ref{fig:timedomain} shows one of the 30 paired trials in the time domain. AA-HalfGain gives a paired RMSE change of $-2.32\pm0.08$ mm relative to AA-Budget, with a modestly higher increment. Among the searched candidates, RO-HalfGain meets the response and sensitivity requirements only with $\omega_o=\omega_c$ in the attitude class (Section~\ref{sec:abl}); it remains bounded but reaches 226.4~mm, $209.5\pm4.8$~mm above AA-HalfGain (all 30 pairs), which supports the role of actuator-input modeling within the half-gain observer family.

\subsection{Comparison under a common strict cascade-increment ceiling}\label{sec:cascade}
\begin{table}[t]\centering\scriptsize
\caption{Cascade Noise: Prediction and Hover Measurement}\label{tab:cascadenoise}
\setlength{\tabcolsep}{4pt}\begin{tabular}{lrrrr}\toprule
Controller & Pred. $\sigma_{\Delta u}$ & Meas. & Ratio $u$ & Ratio $\Delta u$\\\midrule
PID-Budget & 1.380 & 1.382 & 0.998 & 1.003\\
AA-Budget & 0.937 & 0.938 & 0.985 & 1.002\\
AA-HalfGain & 1.023 & 1.024 & 0.982 & 1.001\\
\bottomrule\end{tabular}\par\smallskip\begin{minipage}{\linewidth}\footnotesize Ratios are measured/predicted variances of the inner control $u$ and its increment $\Delta u$; $\sigma_{\Delta u}$ in normalized angular-acceleration units (ten 60-s noise-only hover trials, window 20--60 s).\end{minipage}
\end{table}

\begin{table}[t]
	\centering
	\caption{Cascade Activity Constraints and Performance}
	\label{tab:cascadebudget}
	\scriptsize
	\setlength{\tabcolsep}{2.6pt}
	\renewcommand{\arraystretch}{1.10}
	\begin{tabular}{lccccc}
		\toprule
		\textbf{Design} &
		\textbf{Ceil.} &
		\textbf{Pred.} &
		\textbf{Meas.} &
		\textbf{S6 RMSE} &
		\textbf{Rotor inc.} \\
		&
		\textbf{[rad/s$^2$]} &
		\boldmath$\mathbf{\sigma_{\Delta u}}$ &
		\boldmath$\mathbf{\sigma_{\Delta u}}$ &
		\textbf{[mm]} &
		\textbf{[rad/s]} \\
		\midrule
		PID-Budget    & 1.382 & 1.380 & 1.382 & 65.5$\pm$7.0   & 9.91 \\
		AA-Budget     & 1.382 & 0.937 & 0.938 & 19.2$\pm$1.5   & 7.46 \\
		AA-HalfGain   & 1.382 & 1.023 & 1.024 & \keyresult{16.9$\pm$1.3}   & 7.58 \\
		PID-Budget-S  & 0.598 & 0.592 & 0.592 & 72.3$\pm$7.5   & 9.23 \\
		AA-Budget-S   & 0.598 & 0.496 & \keyresult{0.497} & 29.9$\pm$2.6   & \keyresult{4.67} \\
		AA-HalfGain-S & 0.598 & 0.521 & 0.522 & \keyresult{26.4$\pm$2.3}   & 4.71 \\
		\bottomrule
	\end{tabular}\par\smallskip\begin{minipage}{\linewidth}\footnotesize Measured: larger pooled roll/pitch increment of ten 60-s noise-only trials; tracking: 30 paired S6 trials. The -S designs use the strict ceiling and retune the inner loops.\end{minipage}
\end{table}

The primary cascade threshold, 1.382, is fixed from the PID-Budget roll-control noise-only benchmark rather than selected after observing the ADRC results, so that the reference is common to all designs. Table~\ref{tab:cascadenoise} shows that the full-cascade model predicts the nonlinear hover increments with measured-to-predicted variance ratios of 1.001--1.003: position noise enters the inner loop through the commanded tilt. Because the primary AA designs reach only 0.937 and 1.023, this threshold does not bind them and cannot by itself show the value of the cascade budget. The strict threshold, 0.598, equals the isolated-attitude PID increment and lies below the primary increments of all three designs, which are therefore retuned, including their inner loops. \emph{Under the same strict cascade-level increment ceiling, AA-Budget and AA-HalfGain achieve 29.9 and 26.4~mm, versus 72.3~mm for the retuned PID} (Table~\ref{tab:cascadebudget}; paired changes $-42.4\pm1.9$ and $-45.9\pm2.0$~mm, better in all 30 pairs), and the PID uses roughly double their rotor-command increment (9.23 versus 4.67 and 4.71~rad/s). The ceiling imposes the same upper bound on the predicted roll/pitch cascade increment in noise-only hover, not the total rotor-command activity in disturbed tracking. All strict designs satisfy the local screens, the response requirement and their cascade ceiling, and their measured pooled hover increments stay below it. Relative to the primary designs, the strict AA designs accept a higher RMSE in exchange for lower measured rotor-command increments (4.67 and 4.71 versus 7.46 and 7.58~rad/s). This is evidence for a stationary matched-noise hover constraint, not a cap on rotor-command increments in disturbed tracking.

\subsection{Redesigned ablations}\label{sec:abl}
\begin{table}[t]\centering\scriptsize
\caption{Observer and Budget Ablations (S6, 30 Paired Trials)}\label{tab:ablation}
\setlength{\tabcolsep}{2.5pt}\begin{tabular}{lrrrr}\toprule
Design & RMSE [mm] & CI [mm] & Inc. & Paired change [mm]\\\midrule
A1-Commanded & 176.6$\pm$14.0 & 5.21 & 2.10 & \keyresult{157.4$\pm$4.9}\\
A2-Unaligned & 19.2$\pm$1.5 & 0.58 & 7.70 & 0.0132$\pm$0.0034\\
A3-NoIncrement & 15.9$\pm$1.3 & 0.47 & 8.25 & \keyresult{-3.28$\pm$0.11}\\
A4, fixed gains$^{a}$ & 129.1$\pm$6.9 & 2.56 & 7.84 & 109.9$\pm$2.7\\
A4, reselected & 275.7$\pm$42.2 & 15.75 & 7.16 & \keyresult{256$\pm$16}\\
\bottomrule\end{tabular}\par\smallskip\begin{minipage}{\linewidth}\footnotesize Last column: paired RMSE change versus AA-Budget, mean$\pm$95\% CI. A4: commanded-tilt horizontal observer inputs. $^{a}$Sampled horizontal loop not Schur stable; nonlinear runs stay bounded.\end{minipage}\end{table}

A2 and A3 use twenty logarithmic ratios from 1.5 to 60 under the same local variance, sensitivity and response constraints and the primary cascade ceiling; A3 removes only the local increment ceiling. Inner gains are fixed while the horizontal design is reselected, and all final candidates are audited at full resolution; Table~\ref{tab:ablation} reports the paired outcomes. With commanded inner/altitude inputs, the response and sensitivity requirements are met, within the stated finite search, only at the lowest observer/controller ratios: on the 20-ratio grid from 1.5 to 60 used for the AA designs, A1 is feasible in the attitude class only at the boundary $\omega_o/\omega_c=1.5$, and RO-HalfGain not at all. The ratio range was therefore extended down to $\omega_o=\omega_c$ (four further ratios, 1.0--1.36; lower ratios would no longer give an observer faster than the loop). The best commanded-input designs found for $\omega_o/\omega_c\ge1$, A1 and RO-HalfGain, then reach 176.6 and 226.4~mm, 9.2 times the RMSE of AA-Budget and 13.4 times that of AA-HalfGain, respectively. Both use $\omega_o=\omega_c=3.90$~rad/s in the attitude class. Two sensitivity checks point the same way. The A1 design restricted to $\omega_o/\omega_c\ge1.5$ reaches 245.2~mm (12.8 times; 248.3~mm in PX4 SITL, ten seeds). Restricting the ratio to $\omega_o/\omega_c\in\{0.6,0.8\}$, where the observer is slower than the loop, gives feasible designs that are worse: 363.6 and 389.1~mm for A1 and RO-HalfGain (18.9 and 23.0 times AA-Budget and AA-HalfGain), worse in all 30 pairs. A2 changes the mean by only 0.013 mm at fixed 5-ms measurement age, and the variable-latency tests of Table~\ref{tab:stress} leave the RMSE unchanged without adapting the alignment. At this delay, the benefit of the altitude and attitude input path therefore comes from the rotor-lag model; the effect of alignment at larger nominal delays was not tested. Removing the local increment ceiling in A3 improves RMSE by 17.1\% but increases measured increments by 10.7\%, so the increment constraint regulates a precision/activity trade-off. The ablations combine architecture changes with retuning; their paired differences describe the resulting feasible designs and do not isolate the input change at fixed gains, which in the nonlinear runs was done only for the horizontal path (A4). A4-CommandedTilt replaces the measured-attitude input of the horizontal observers by the commanded tilt and keeps the actuator-aware altitude and attitude observers. With the AA-Budget gains its sampled horizontal loop is not Schur stable, although the 30 nonlinear runs stay bounded at 129.1~mm; the best feasible horizontal design found on the stated grid, $(\omega_c,\omega_o)=(0.41,2.39)$~rad/s, reaches 275.7~mm, 14.4 times AA-Budget, worse in all 30 pairs (S5: 274.5~mm). Both input paths therefore materially improve the best-found tracking under the tested design procedure; replacing the measured-attitude input by the commanded tilt degrades the S6 RMSE to 275.7~mm after horizontal reselection.

\subsection{Local budget--performance trade-off}
\begin{figure*}[tbp]\centering\includegraphics[width=.84\textwidth,height=.18\textheight,keepaspectratio]{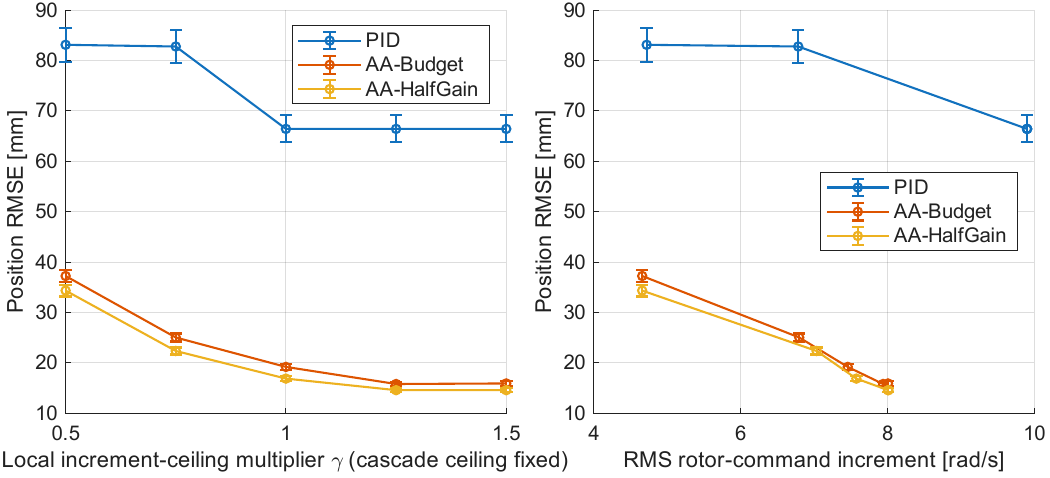}\caption{Local budget--performance trade-off of PID, AA-Budget and AA-HalfGain redesigned for local increment-ceiling multipliers $\gamma\in\{0.5,0.75,1,1.25,1.5\}$ with the cascade ceiling fixed. S6 position RMSE (mean and 95\% CI over 30 paired trials) versus $\gamma$ (left) and versus the measured rotor-command increment (right).}\label{fig:budget}\end{figure*}

For $\gamma\in\{0.5,0.75,1,1.25,1.5\}$, PID, AA-Budget and AA-HalfGain candidates are redesigned with scaled local increment ceilings and compared on the same thirty exogenous realizations (Fig.~\ref{fig:budget}). Only the local increment ceilings are scaled by $\gamma$; the other constraints stay fixed. The left panel is plotted against the local multiplier and the right panel against the measured rotor-command increment, with 95\% CIs over the 30 pairs. The PID points at $\gamma=0.5$ and 0.75 are redesigned under the response requirement (Appendix~\ref{app:search}); at $\gamma=0.75$ the best PID found keeps the attitude, horizontal and yaw gains of $\gamma=0.5$. Each search includes its local-budget incumbent and a finite ratio/bandwidth scan subject to the common cascade cap. Candidate lists are not nested, so small nonmonotonic changes are possible; the plateau represents the best candidates found by the stated procedure. The redesigned PID reaches 83.1 and 82.8~mm at $\gamma=0.5$ and 0.75 and 66.4~mm for $\gamma\ge1$. The latter point is the budget-curve candidate of the PID search. It has the inner, altitude and yaw gains of PID-Budget, but its horizontal gains were reselected by the curve search ($K_p,K_i,K_d,\omega_f$ = 4.395, 3.279, 3.190, 15242 versus 4.465, 3.358, 3.215, 15363 for PID-Budget). This is why it reaches 66.4 rather than the 65.5~mm of PID-Budget in Table~\ref{tab:nominal} (0.98~mm higher in all 30 pairs). Both AA families are better than PID in all 30 pairs at every level (e.g., 37.2 and 34.3~mm at $\gamma=0.5$).

\subsection{Noise and variable measurement age}
\begin{table*}[t]
	\centering
	\caption{Noise and Latency Stress Tests (S6, Ten Paired Trials per Condition)}
	\label{tab:stress}
	\scriptsize
	\setlength{\tabcolsep}{5.5pt}
	\renewcommand{\arraystretch}{1.12}
	\begin{tabular}{lcccc}
		\toprule
		\textbf{Condition} &
		\textbf{PID-Budget} &
		\textbf{AA-Budget} &
		\textbf{RO-HalfGain} &
		\textbf{AA-HalfGain} \\
		&
		\multicolumn{4}{c}{\textbf{RMSE [mm] [rotor-command increment, rad/s]}} \\
		\midrule
		Reference
		& 68.3$\pm$7.6 [9.91]
		& 19.6$\pm$1.3 [7.45]
		& 227.4$\pm$19.4 [0.79]
		& \keyresult{17.2$\pm$1.1} [7.57] \\
		
		Noise $0.5\times$
		& 68.4$\pm$7.6 [4.96]
		& 19.6$\pm$1.3 [3.73]
		& 227.5$\pm$19.4 [0.40]
		& \keyresult{17.2$\pm$1.1} [3.79] \\
		
		Noise $2\times$
		& 68.3$\pm$7.6 [19.84]
		& 19.7$\pm$1.3 [14.92]
		& 227.3$\pm$19.3 [1.59]
		& \keyresult{17.3$\pm$1.1} [15.15] \\
		
		Noise $4\times$
		& 68.3$\pm$7.7 [39.76]
		& 20.0$\pm$1.2 [29.91]
		& 227.2$\pm$19.1 [3.18]
		& \keyresult{17.7$\pm$1.0} [30.39] \\
		
		Delay 0--10 ms
		& 68.3$\pm$7.6 [9.86]
		& 19.6$\pm$1.3 [7.42]
		& 227.8$\pm$19.4 [0.79]
		& \keyresult{17.2$\pm$1.1} [7.53] \\
		
		Delay bursts 20 ms
		& 68.3$\pm$7.6 [9.91]
		& 19.6$\pm$1.3 [7.45]
		& 228.3$\pm$19.1 [0.80]
		& \keyresult{17.2$\pm$1.1} [7.57] \\
		\bottomrule
	\end{tabular}\par\smallskip\begin{minipage}{\linewidth}\footnotesize RMSE mean$\pm$SD; brackets: rotor-command increment.\end{minipage}
\end{table*}

\begin{figure*}[tbp]\centering\includegraphics[width=.84\textwidth,height=.2\textheight,keepaspectratio]{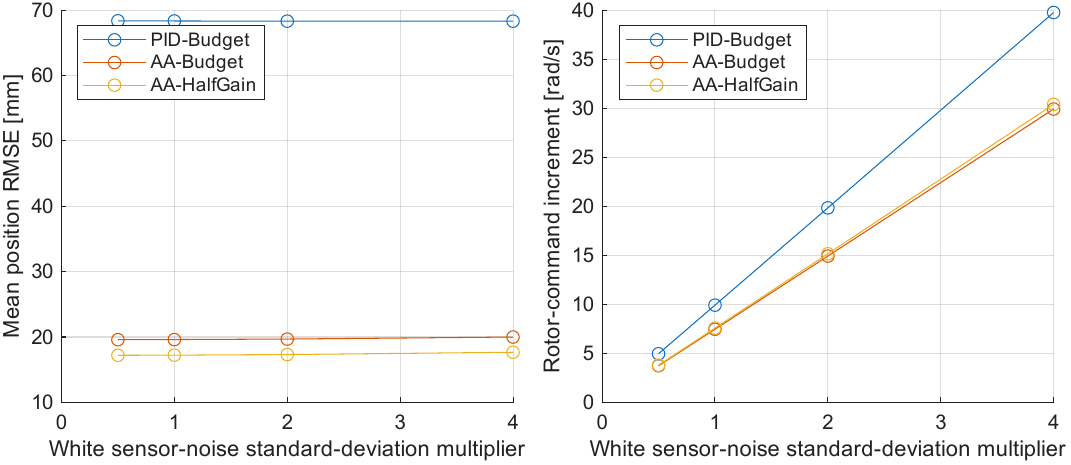}\caption{Effect of scaling the white sensor-noise standard deviations by 0.5--4 in S6 with fixed gains (ten paired trials per condition). Left: mean position RMSE. Right: measured rotor-command increment.}\label{fig:noise}\end{figure*}

Table~\ref{tab:stress} and Fig.~\ref{fig:noise} separate tracking error from the measured rotor-command increment. White sensor-noise standard deviations are scaled together; gyro-bias walk, wind, gains and internal lag/alignment remain fixed. Variable measurement age is either $n_d(k)=\operatorname{round}[5+5\sin(2\pi kT_s)]$ or fixed 5 ms with 200-ms bursts to 20 ms every 2 s from $t=10$ s; a 31-column history selects the noisy sample at its acquisition timestamp. These tests record 0 failures in 240 realizations. Increased noise changes tracking much less than command increments, exposing activity costs concealed by RMSE alone. Both latency patterns are evaluated without adapting the measurement-time alignment.

\subsection{Fixed-gain mismatch as a limitation}
\begin{figure*}[tbp]\centering\includegraphics[width=.84\textwidth,height=.2\textheight,keepaspectratio]{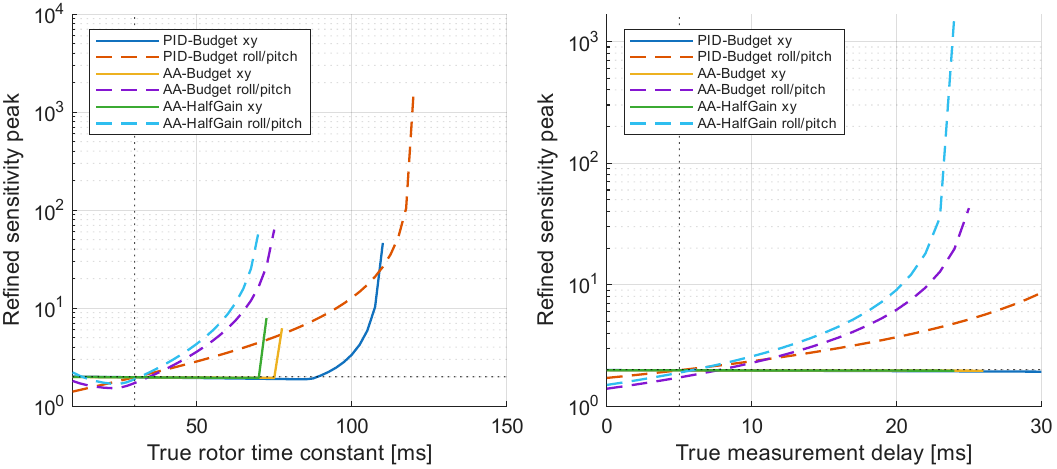}\caption{Sensitivity peak of the horizontal ($xy$, solid) and roll/pitch (dashed) loops of PID-Budget, AA-Budget and AA-HalfGain with fixed nominal gains under plant mismatch. Left: true rotor time constant at the 5-ms delay. Right: true measurement delay at the 30-ms time constant. Vertical dotted lines: nominal values; horizontal dotted line: $M_s=2$; gaps: loops that are not Schur stable.}\label{fig:mismatchms}\end{figure*}

\begin{figure*}[tbp]\centering\includegraphics[width=.82\textwidth]{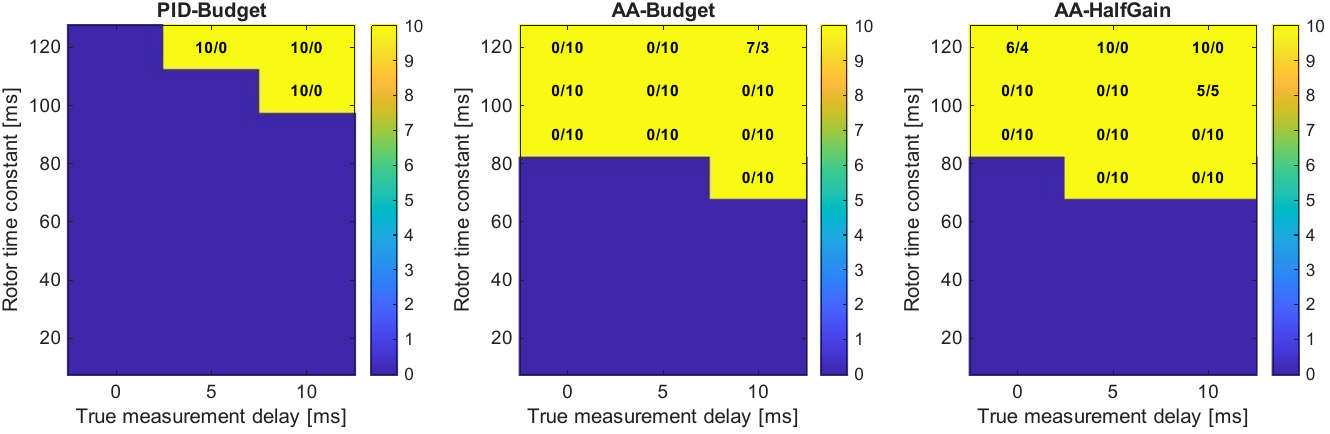}\caption{Per-plant outcomes of PID-Budget, AA-Budget and AA-HalfGain with fixed nominal gains for each true rotor time constant and measurement delay. Colour: failures plus bounded rotor-fluctuation threshold exceedances out of ten paired S6 trials; the cell labels F/E give the two disjoint counts.}\label{fig:mismatchcounts}\end{figure*}

Nominal gains and internal $(\hat\tau_m,\hat n_d)=(30\mathrm{\ ms},5)$ are frozen. Linear screens sweep true lag 10--150 ms in 2.5-ms steps and integer delay 0--30 ms. Nonlinear tests use lags $\{15,30,45,60,75,90,105,120\}$ ms and delays $\{0,5,10\}$ ms, ten paired seeds per plant. Figure~\ref{fig:mismatchms} reports the horizontal and roll/pitch sensitivity screens: the left panel varies the true rotor time constant at the 5-ms delay and the right panel the true delay at the 30-ms time constant; gaps are non-Schur loops and the horizontal line marks $M_s=2$. Table~\ref{tab:mismatch} aggregates counts over twelve plants per range and controller, and Fig.~\ref{fig:mismatchcounts} resolves the individual plants: each cell gives, out of ten paired S6 trials, the failures F and the bounded threshold exceedances E (disjoint counts).

\begin{table}[t]\centering\scriptsize
\caption{Fixed-Gain Mismatch Outcomes (Twelve Plants per Lag Range, Ten Paired S6 Trials Each)}\label{tab:mismatch}
\setlength{\tabcolsep}{4pt}\begin{tabular}{lrrrr}\toprule
Controller & Lag [ms] & Runs & Fail & Threshold\\\midrule
PID-Budget & 15--60 & 120 & 0 & 0\\
PID-Budget & 75--120 & 120 & 30 & 0\\
AA-Budget & 15--60 & 120 & 0 & 0\\
AA-Budget & 75--120 & 120 & \keyresult{7} & \keyresult{93}\\
AA-HalfGain & 15--60 & 120 & 0 & 0\\
AA-HalfGain & 75--120 & 120 & 31 & 79\\
\bottomrule\end{tabular}\par\smallskip\begin{minipage}{\linewidth}\footnotesize Threshold: bounded runs with rotor fluctuation above 20 rad/s; failures counted separately.\end{minipage}
\end{table}

The two actuator-aware variants remain bounded with no threshold exceedance in the tested 15--60-ms plants. In the 75--120-ms range, AA-Budget has 7 failures and 93 bounded threshold exceedances; AA-HalfGain has 31 failures and 79 exceedances. AA-HalfGain has the lowest mean RMSE among the four primary designs in S6 and S5, but AA-Budget has fewer failures under the tested large rotor-lag mismatch, so the nominal advantage of the tuned half-gain design does not imply uniformly improved robustness. PID-Budget records 30 failures and 0 bounded threshold exceedances in the outer lag range; its sampled linear margin is larger over part of the grid, yet nonlinear failures occur at the highest lag/delay combinations. In the linear screen of Fig.~\ref{fig:mismatchms}, the roll/pitch loops of both AA designs lose Schur stability at a true delay of roughly 23--25~ms and a true rotor time constant of roughly 70--75~ms, whereas the PID roll/pitch loop remains Schur stable up to 30~ms of delay and to about 120~ms of time constant. Because every design is tuned close to the $M_s$ limit, the roll/pitch $M_s$ of all designs exceeds 2 already for small increases of the true delay or lag above the nominal values. The nominal advantage of the AA designs is thus paid for with a smaller tolerance to unmodelled lag and delay. Robust redesign over an uncertainty set is the subject of a companion study.

\subsection{Paired lemniscate tracking}
\begin{table}[t]
	\centering
	\caption{Lemniscate Tracking (S5, 30 Paired Trials, Window 5--45 s)}
	\label{tab:s5}
	\scriptsize
	\setlength{\tabcolsep}{3.2pt}
	\renewcommand{\arraystretch}{1.10}
	\begin{tabular}{lccccc}
		\toprule
		\textbf{Controller} &
		\textbf{RMSE} &
		\textbf{95\% CI} &
		\textbf{Inc.} &
		\textbf{Max. error} &
		\textbf{Fail} \\
		&
		\textbf{[mm]} &
		\textbf{[mm]} &
		\textbf{[rad/s]} &
		\textbf{[mm]} &
		\\
		\midrule
		PID-Budget   & 276.5$\pm$7.7    & 2.88  & 9.92 & 464.0$\pm$32.0   & 0/30 \\
		AA-Budget    & 22.7$\pm$1.7     & 0.64  & 7.46 & 94.9$\pm$14.5    & 0/30 \\
		RO-HalfGain  & 217.3$\pm$16.2   & 6.04  & 0.79 & 641.2$\pm$86.6   & 0/30 \\
		AA-HalfGain  & \keyresult{20.4$\pm$1.5}     & 0.55  & 7.58 & \keyresult{91.6$\pm$13.2}    & 0/30 \\
		\bottomrule
	\end{tabular}\par\smallskip\begin{minipage}{\linewidth}\footnotesize Mean$\pm$SD; maximum error: per-trial maximum 3-D error norm.\end{minipage}
\end{table}

Table~\ref{tab:s5} reports the trajectory statistics. The horizontal reference starts at $t=5$ s with a velocity step from zero to $\pi/10$ m/s on both axes, with unchanged velocity and acceleration feed-forward for every controller. Because the S5 metric window starts at 5~s, its maximum error includes the transient caused by this velocity jump. AA-Budget and AA-HalfGain keep the set-point ranking on the moving reference (22.7 and 20.4~mm versus 276.5~mm for PID-Budget). The PID error on this reference depends on the set-point weight: with the searched weights of $xy$ and $\psi$ before the rule of Section~\ref{sec:req}, PID-Budget reaches 339.0~mm in S5 for an unchanged S6 RMSE, because the term $K_p(1-b)\chi_d$ acts as a reference-proportional disturbance that the integrator must remove. The weights follow the same declared rule for all PID designs, so the S5 result characterizes PID designs selected under the common response specification, not the best trajectory tracking attainable with a PID structure.

\subsection{PX4 software-in-the-loop check}\label{sec:sitl}
The designs were also run in PX4~\cite{meier2015px4} v1.16.1 software-in-the-loop (SITL). The controller functions were code-generated with the UAV Toolbox Support Package for PX4 and executed every 1~ms inside PX4 (its multicopter controllers disabled). Gains and internal models were unchanged, but the controllers were fed by the EKF2 position, velocity and attitude estimates and the filtered gyroscope rate. The plant of Section~\ref{sec:model}, with the wind, torque and gyro-bias sequences of the same seeds, runs in lockstep with PX4 over MAVLink. It supplies 1-kHz inertial samples (gyroscope as in Table~\ref{tab:params}, accelerometer noise 0.05~m/s$^2$) and 100-Hz motion-capture-like pose and velocity samples with the noise of Table~\ref{tab:params} and a 5-ms latency. PX4 returns actuator outputs quantized to 1/1000 of full scale (1~rad/s in $\wcmd$). The 1-ms lockstep execution checks the generated code and the estimator interface; it is not evidence of execution time on flight hardware.

\begin{table}[t]\centering\scriptsize
\caption{MATLAB and PX4 SITL Position RMSE (S6, 10--40~s; S5, 5--45~s)}\label{tab:sitl}
\setlength{\tabcolsep}{3pt}\begin{tabular}{llrr}\toprule
Case & Design & MATLAB & PX4 SITL\\\midrule
\multirow{4}{*}{S6} & PID-Budget & 65.5$\pm$7.0 & 65.0$\pm$6.7\\
 & AA-Budget & 19.2$\pm$1.5 & 20.5$\pm$1.3\\
 & RO-HalfGain & 226.4$\pm$13.4 & 227.1$\pm$12.9\\
 & AA-HalfGain & 16.9$\pm$1.3 & 18.3$\pm$1.3\\\midrule
\multirow{3}{*}{S6 strict} & PID-Budget-S & 72.3$\pm$7.5 & 72.2$\pm$7.2\\
 & AA-Budget-S & 29.9$\pm$2.6 & 30.8$\pm$2.6\\
 & AA-HalfGain-S & 26.4$\pm$2.3 & 27.4$\pm$2.2\\\midrule
\multirow{4}{*}{S5} & PID-Budget & 279.4$\pm$7.8 & 278.4$\pm$8.3\\
 & AA-Budget & 22.5$\pm$2.4 & 24.2$\pm$2.5\\
 & RO-HalfGain & 214.1$\pm$19.5 & 215.8$\pm$18.0\\
 & AA-HalfGain & 20.2$\pm$2.0 & 21.8$\pm$2.3\\\midrule
\multirow{3}{*}{S6 ablation} & A1-Commanded & 178.2$\pm$15.3 & 178.8$\pm$15.1\\
 & A2-Unaligned & 19.5$\pm$1.4 & 20.5$\pm$1.3\\
 & A3-NoIncrement & 16.1$\pm$1.2 & 17.4$\pm$1.2\\
\bottomrule\end{tabular}\par\smallskip\begin{minipage}{\linewidth}\footnotesize RMSE in mm, mean$\pm$SD on the same seeds (S6: 91001--91030; S5 and ablations: 91001--91010). RO-HalfGain and A1: $\omega_o/\omega_c\ge1$; PID-Budget: set-point weights of Section~\ref{sec:req}. No failures.\end{minipage}\end{table}

(i) \emph{Tracking.} Table~\ref{tab:sitl} compares the SITL runs with the MATLAB results of the same seeds. In S6, AA-Budget improves on PID-Budget by $44.5\pm2.2$~mm and AA-HalfGain on AA-Budget by $2.22\pm0.27$~mm. Under the strict ceiling, AA-Budget-S and AA-HalfGain-S reach 30.8 and 27.4~mm against 72.2~mm for PID-Budget-S ($-41.4\pm1.8$ and $-44.8\pm2.0$~mm; better in all 30 pairs in each case). The commanded-input designs stay far above the actuator-aware ones (A1 178.8~mm, RO-HalfGain 227.1 and 215.8~mm in S6 and S5, worse than AA-Budget or AA-HalfGain in every pair), and the ablation ordering is unchanged. With true rotor lags of 75--120~ms ($n_d=5$, seeds 92001--92003; four lags $\times$ three seeds for each of the three primary designs, 36 runs), the failure and fluctuation-threshold counts coincide with MATLAB: only PID-Budget and AA-HalfGain fail, at 120~ms. The actuator-aware designs lose 0.9--1.4~mm in S6 because they follow the estimate, whose position error in AA-Budget runs has an RMS of 3.7~mm; only the filtered rates show a material lag (see (ii)). The lockstep loop is not bit-reproducible (three AA-Budget runs of one seed: 20.1--21.8~mm), and PX4 occasionally skipped the output of a step: in 41 runs outside Table~\ref{tab:miss} (one or two steps each) and in 101 of the 390 runs listed there (at most nine steps per run). Removing the pairs with skipped steps changes no paired conclusion (Appendix~\ref{app:audit}).

(ii) \emph{Activity.} The white-noise ceilings do not describe the estimator-fed loop. Logged inside PX4 before output quantization (ten paired S6 seeds), the rotor-command increments are 0.084, 0.183 and 0.153~rad/s for PID-Budget, AA-Budget and AA-HalfGain (MATLAB, same seeds: 9.91, 7.45 and 7.57~rad/s), so the activity ordering reverses. When the estimator is bypassed and the same code reads the measurement model of Table~\ref{tab:params} at 1~kHz with a 5-ms delay, the increments return to 9.91, 7.46 and 7.57~rad/s. The hover increments $\sigma_{\Delta u}$ of the five designs of Table~\ref{tab:cascadebudget} whose gains are unchanged then agree with it within 0.3\%. The estimator therefore causes the reversal; code generation, lockstep timing and quantization do not. For the model of Section~\ref{sec:estaware}, the controller logs the twelve signals it uses at every step and the plant state gives their errors, in 60 noise-only hover runs (six designs, including a superseded strict PID design whose logs serve only as fitting data; seeds 96001--96010, window 20--60~s). The EKF2 errors are concentrated below about 1~Hz, whereas the rate errors carry the gyroscope noise through the 40-Hz PX4 filter. Only the rates show a material lag (6~ms, which reduces their S6 error RMS to about 0.3 of the zero-lag value), so they keep a 6-ms buffer and the other signals none. For each channel class a vector autoregressive model of the error vector, fitted on seeds 96001--96005 (orders 13--15 except one first-order yaw model, by the Bayesian information criterion among stationary fits; one model per design, because the attitude-error increments differ up to fivefold between designs), gives \eqref{eq:shaping}. The error statistics therefore depend on the closed loop; at design time the model pooled over the designs is used (validated below). On the held-out seeds the standard deviations of its error increments agree within 2\% for the rate errors and within 10\% for the other outputs except one position channel (20\%), whereas the variance of the slowly drifting attitude errors is not reproduced. With the 6-ms rate delay the input-sensitivity peak becomes 2.04, 1.92 and 2.02 for PID-Budget, AA-Budget and AA-HalfGain (1.97--1.98 without), mainly in the attitude class.

\emph{Validation at three levels.} The 60-s logs are short compared with the slow EKF2 errors, so the model was validated on 300-s noise-only hover runs (window 20--300~s) of nine designs: the three primary, the three strict and the three -E designs of (iii). Seeds 97001--97005 were used for fitting and 97006--97010 were held out. The three levels are (a) one model per design fitted on its own 300-s logs; (b) one model pooled over the nine designs; and (c) for the -E designs, the pooled model of the design stage, fitted on the 60-s logs above before any -E run (refitted with the 6-ms delay on the yaw rate as well, Appendix~\ref{app:search}). Table~\ref{tab:estaware} lists the predicted-to-measured ratios of the standard deviations. The model predicts the roll/pitch and yaw increments within 5\% on held-out and out-of-sample runs, and the total rotor-command increment within 14\%. It is less accurate for the variances. The control standard deviation is underpredicted by up to 19\%, and the tracking standard deviation is matched within about 25\% except the out-of-sample yaw tracking error (about 60\%, see below). The variance objective is therefore an approximate design criterion. The pooled model underpredicts the altitude increment of the three PID designs by 17--19\% and their $J$ by up to 14\%, because the altitude estimation error depends on the controller; the per-design models stay within 6\%. The out-of-sample yaw tracking error is overpredicted by about 60\% at level (c) but matched within 5\% at level (b): its low-frequency spectrum cannot be identified from the 40-s window of the design-stage logs. The underprediction of $\sigma_u$ is non-conservative but does not affect feasibility, since the measured control standard deviations of the -E designs stay 10--20 times below the local ceilings $\bar u_c$. On the same held-out runs, the white-noise model of the MATLAB designs overpredicts the roll/pitch inner increments by factors of 19 to 217.

\begin{table}[t]\centering\scriptsize
\caption{Noise-Model Validation Behind EKF2 (300-s Noise-Only Hover, Window 20--300~s, Nine Designs)}\label{tab:estaware}
\setlength{\tabcolsep}{2pt}\begin{tabular}{lccccc}\toprule
 & \multicolumn{2}{c}{(a) per design} & \multicolumn{2}{c}{(b) pooled} & (c) design stage\\\cmidrule(lr){2-3}\cmidrule(lr){4-5}\cmidrule(lr){6-6}
Quantity & fit & h.o. & fit & h.o. & h.o.\\\midrule
$\sigma_{\Delta u}$, $\phi\theta$ & 0.98--1.00 & 0.98--1.00 & 0.96--1.01 & 0.95--1.01 & 0.97--1.01\\
$\sigma_{\Delta u}$, $\psi$ & 1.00 & 1.00--1.01 & 0.99--1.00 & 1.00--1.01 & 1.01--1.02\\
$\sigma_{\Delta u}$, $z$ & 0.98--1.06 & 0.96--1.06 & 0.83--1.07 & 0.81--1.07 & 0.83--1.07\\
$J$ & 0.98--1.03 & 0.97--1.03 & 0.87--1.04 & 0.86--1.04 & 0.93--1.04\\
$\sigma_u$, $\phi\theta$ & 0.81--0.97 & 0.81--0.97 & 0.81--0.97 & 0.81--0.97 & 0.87--0.93\\
$\sigma_u$, $z$ & 0.94--1.04 & 0.94--1.02 & 0.93--1.04 & 0.93--1.02 & 0.93--1.01\\
$\sigma_u$, $\psi$ & 0.83--1.02 & 0.84--1.02 & 0.83--1.02 & 0.84--1.02 & 0.91--1.02\\
$\sigma_e$, $x,y$ & 0.95--1.22 & 0.93--1.21 & 0.94--1.25 & 0.95--1.23 & 0.94--1.20\\
$\sigma_e$, $z$ & 0.91--1.08 & 0.87--1.07 & 0.89--1.08 & 0.89--1.07 & 0.95--1.07\\
$\sigma_e$, $\psi$ & 0.97--1.01 & 0.89--1.06 & 0.94--1.03 & 0.91--1.05 & 1.58--1.61\\
\bottomrule\end{tabular}\par\smallskip\begin{minipage}{\linewidth}\footnotesize Predicted/measured SD, min--max over designs; fit seeds 97001--97005, held-out (h.o.) 97006--97010. (a) Per-design model; (b) pooled model; (c) design-stage model of the -E designs (out of sample). $\sigma_{\Delta u}$: inner increment; $\sigma_u$: control; $\sigma_e$: tracking error; $J$: rotor-command increment.\end{minipage}\end{table}

\begin{table}[t]\centering\scriptsize
\caption{Estimator-Aware Designs: Sensitivity and Hover (EKF2, Ten 60-s Runs, Seeds 96001--96010)}\label{tab:equalE}
\setlength{\tabcolsep}{1.6pt}\begin{tabular}{lccccc}\toprule
Design & $M_s$ & $\sigma_{\Delta u}$ cas. & $\sigma_{\Delta u_\psi}$ & $\sigma_{\Delta u_z}$ & $J$\\
 & $xy/z/\phi\theta/\psi$ & pred./meas. & pred./meas. & meas. & pred./meas.\\\midrule
PID-Budget & 1.94/1.81/2.04/1.99 & 0.0065/0.0066 & 4.75/4.74 & 5.0 & --/0.0259\\
PID-E & 1.98/1.81/1.98/1.98 & 0.0064/0.0067 & 4.22/4.21 & 4.8 & 0.0236/0.0254\\
AA-Budget-E & 1.98/1.46/1.96/1.97 & 0.0059/0.0059 & 4.31/4.24 & 7.0 & 0.0302/0.0290\\
AA-HalfGain-E & 1.98/1.58/1.81/1.97 & 0.0064/0.0064 & 3.99/3.93 & 7.4 & 0.0317/0.0305\\
\bottomrule\end{tabular}\par\smallskip\begin{minipage}{\linewidth}\footnotesize $M_s$ with the 6-ms rate lag; ceilings $B^{E}_{\mathrm{cas}}=0.0065$, $B^{E}_{\psi}=4.75\times10^{-4}$; predictions out of sample; $\sigma_{\Delta u_\psi}$, $\sigma_{\Delta u_z}$ in $10^{-4}$, $J$ in rad/s. PID-Budget is not redesigned.\end{minipage}\end{table}

\begin{table*}[t]\centering\scriptsize
\caption{Estimator-Aware Designs in PX4 SITL (S6: 30 Paired Trials; S5: Ten)}\label{tab:equalEb}
\setlength{\tabcolsep}{4pt}\begin{tabular}{lcccccccc}\toprule
 & \multicolumn{5}{c}{S6} & \multicolumn{3}{c}{S5}\\\cmidrule(lr){2-6}\cmidrule(lr){7-9}
Design & RMSE & vs PID-B. & vs PID-E & $J$ & $J_q$ & RMSE & $J$ & $J_q$\\
 & [mm] & [mm] & [mm] & [rad/s] & [rad/s] & [mm] & [rad/s] & [rad/s]\\\midrule
PID-Budget & 65.0$\pm$6.7 & -- & -- & 0.084 & 0.224 & 278.4$\pm$8.3 & 0.222 & 0.273\\
PID-E & 59.1$\pm$6.3 & $-5.9\pm1.0$ (30) & -- & 0.089 & 0.231 & 335.8$\pm$6.6 & 0.228 & 0.294\\
AA-Budget-E & \keyresult{44.0$\pm$4.0} & $-21.0\pm1.3$ (30) & $-15.0\pm1.1$ (30) & \keyresult{0.049} & 0.196 & \keyresult{49.5$\pm$5.6} & 0.111 & 0.205\\
AA-HalfGain-E & 84.7$\pm$8.2 & $+19.7\pm1.0$ (0) & $+25.6\pm1.1$ (0) & 0.047 & 0.192 & 90.6$\pm$11.9 & 0.097 & 0.194\\
\bottomrule\end{tabular}\par\smallskip\begin{minipage}{\linewidth}\footnotesize RMSE mean$\pm$SD; paired change $\pm$95\% CI (pairs better); $J$ before and $J_q$ after the 1/1000 output quantizer. PID-Budget: set-point weights of Section~\ref{sec:req}, same build and seeds as the -E runs. No failures.\end{minipage}\end{table*}

(iii) \emph{Redesign for the estimator-fed loop.} With the 6-ms rate lag, the input-sensitivity peaks of PID-Budget and AA-HalfGain are 2.04 and 2.02 (attitude class), above the specification, whereas AA-Budget stays at 1.92. AA-Budget and AA-HalfGain, however, exceed the estimator-aware cascade ceiling by factors of 5.5 and 3.6 and the yaw ceiling by factors of 15.7 and 13.9. For a like-for-like comparison all three were redesigned by the same procedure. PID-E, AA-Budget-E and AA-HalfGain-E were therefore designed by the procedure of Section~\ref{sec:estaware} with the frozen ceilings $B^{E}_{\mathrm{cas}}$ and $B^{E}_{\psi}$. All three meet $M_s\le1.98$ in every class with the 6-ms lag, the response requirement, the local ceilings and, in the design-stage prediction, both estimator-aware ceilings (gains in Table~\ref{tab:gains2}; the altitude classes of PID-Budget and AA-Budget already met them and were kept). After their hover logs, the refitted pooled model left every constraint satisfied (predicted-to-measured roll/pitch cascade increment 0.964, 0.998 and 1.008, yaw 0.998, 1.008 and 1.008), so the gains were frozen after the first design. The design-stage predictions of Table~\ref{tab:equalE} are therefore out of sample: the roll/pitch and yaw increments are matched within 4\% and 2\%, whereas $J$ is underpredicted by 7\% for PID-E and overpredicted by 4\% for the AA designs, mainly through the altitude class. In S6 (Table~\ref{tab:equalEb}), AA-Budget-E reduces the RMSE from 59.1~mm (PID-E) to 44.0~mm, better in all 30 pairs. Its rotor-command increment is lower than that of PID-E and PID-Budget in all 30 S6 and 10 S5 pairs, both before and after output quantization. The mean reductions relative to PID-E are 45\% and 15\% in S6 and 51\% and 30\% in S5. AA-HalfGain-E is better on the lemniscate but worse on the set-point scenario (84.7~mm, worse than PID-E in all 30 pairs), with a lower increment than both PID designs in all pairs. Its larger tracking variance was already apparent at design time (predicted horizontal $\sigma_e$ of 183~mm versus 83 and 108~mm for AA-Budget-E and PID-E). In S5 both AA designs stay far below the PID designs (49.5 and 90.6 versus 335.8 and 278.4~mm). In S5, PID-E is worse than PID-Budget by $57.3\pm2.6$~mm (all ten pairs), whereas it is better in S6 ($-5.9\pm1.0$~mm, all 30 pairs): for the selected PID gains its set-point weights stay of order $10^{-3}$ because the overshoot limit is active (Section~\ref{sec:req}). The pooled measured hover cascade increment of PID-E is 0.00666, 2.3\% above the frozen ceiling (full precision 0.0065137; 1.5\% in the independent rerun, Section~\ref{sec:numver}), within the 4--5\% validation error; the mean of its per-seed maximum-axis increments is $0.00673\pm0.00009$ (95\% CI, ten seeds). The unchanged PID-Budget, which defines the ceiling through its prediction, measures 0.00660 (1.3\% above; per-seed mean $0.00664\pm0.00007$), and AA-Budget-E and AA-HalfGain-E stay 8.7 and 2.0\% below it. In noise-only hover, $J$ of AA-Budget-E and AA-HalfGain-E is 12 and 18\% above that of PID-Budget (0.0290 and 0.0305 versus 0.0259~rad/s; 15 and 20\% in the rerun), because the altitude class has no ceiling ($\sigma_{\Delta u_z}=7.0$ and $7.4$ versus $5.0\times10^{-4}$); PID-E is 2\% below it.

\subsection{Numerical verification}\label{sec:numver}
All fixed protocols were rerun from a fresh copy of the code with the frozen full-precision gains, without re-optimization or noise-model refitting. In MATLAB, all 3310 simulation calls reproduce the reported metrics within a relative difference of $10^{-14}$, with identical failure classifications. In PX4 SITL, where the lockstep loop is not deterministic, the mean RMSEs of Tables~\ref{tab:sitl} and \ref{tab:equalEb} change by at most 1.14~mm (PID-Budget, S5: 278.44 and 279.58~mm), and every count of better pairs is unchanged; AA-Budget-E improves on PID-E by $-14.7\pm0.9$~mm in S6 (reported campaign: $-15.0\pm1.1$~mm). The tables report the first campaign, whose run plans were fixed beforehand; the rerun serves only as a reproducibility check. Appendix~\ref{app:audit} gives the remaining checks.

\section{Discussion and Scope}\label{sec:discussion}
\emph{Interpretation.} The evidence supports actuator-aware observer inputs as a nominal design choice for the tested cascade rather than a universally necessary architecture. Budget and half-gain tuning are complementary variants, not a ranking. In MATLAB, the half-gain variant is better at set-point and on the lemniscate, whereas on the estimator stack only AA-Budget-E beats PID in both scenarios.

\emph{Model and assumptions.} The covariances are exact only for the implemented linear sampled model. The plant relies on assumed sensor, rotor and wind specifications and Gaussian noise. Frame vibration, propeller-induced coloured inertial-measurement-unit (IMU) noise and asymmetric motor spin-up and spin-down are not modelled. It uses Euler-angle kinematics without yaw-angle wrapping (unnecessary for the bounded yaw references used) and componentwise allocation clipping, without ground contact. Consequently, the stationary cascade ceiling closes the modeled position-to-attitude noise path at nominal hover. It does not bound nonlinear rotor-command activity under mismatch, varying noise or clipping, and command increments remain surrogates for motor power, heating or lifetime; whether the 15\% reduction after output quantization in Section~\ref{sec:sitl} matters for motor heating or wear was not assessed. With the 6-ms filtered-rate lag of the PX4 stack, the sampled input-sensitivity peak of the nominal PID-Budget and AA-HalfGain rises to 2.04 and 2.02, slightly above the reporting limit. The -E designs include this lag in the $M_s$ screen and meet 1.98 in every class.

\emph{Estimator-coloured noise model.} The identified noise model depends on the controller as well. For the -E designs, the redesign--refit check confirmed all constraints after the first design. However, the pooled model is less accurate for the altitude class of the PID designs and for the control and tracking standard deviations, including an out-of-sample yaw tracking error about 60\% too large (Table~\ref{tab:estaware}). The variance objective is therefore approximate. The model is specific to the EKF2, sensor, rate-filter and delay configuration of this PX4 stack. It must be re-identified before its predictions are reused with another estimator, sensor set, tuning or autopilot configuration.

\emph{Identification and robustness.} The AA designs rely on the internal $(\hat\tau_m,\hat n_d)$, and Fig.~\ref{fig:mismatchms} shows that their tolerance to unmodelled lag and delay is smaller than that of PID. On a vehicle, $\hat\tau_m$ should therefore be identified, for example from motor-speed or thrust step responses on a bench, and $\hat n_d$ from time-stamped logs of the sensor and control paths, including changes of $\tau_m$ with battery voltage, which this study does not model.

\emph{Dependence on the design space.} The lemniscate error of the PID designs depends on the set-point weight, which the common overshoot limit keeps small for PID-Budget-S and PID-E. The horizontal derivative filter of PID-Budget has $\omega_f$ far above $\pi/T_s\approx3142$~rad/s, so $a_f\approx0$ and the derivative term uses the unfiltered rate of the previous sample; this is an outcome of the search, and a PID with a reference prefilter instead of a set-point weight was not examined. With commanded inputs, the response and sensitivity requirements are met only at the lowest searched observer/controller ratios. The magnitude of the commanded-input ablation therefore depends on the admissible ratio range: 9.2-fold for $\omega_o/\omega_c\ge1$, 12.8-fold for $\omega_o/\omega_c\ge1.5$ and 18.9-fold for $\omega_o/\omega_c\in\{0.6,0.8\}$.

\emph{Finite checks.} Frequency, ratio and uncertainty grids and finite paired samples do not certify continuous-frequency or nonlinear stability or global design optimality. The paired intervals are per comparison and not adjusted for multiplicity. The comparisons on which the conclusions rest differ from zero by many times their interval half-widths; the smallest reported differences (0.013~mm for A2 and 0.98~mm between PID-Budget and its budget-curve counterpart) are practically negligible whether or not they are statistically significant after adjustment. A rotor-fluctuation exceedance alone does not prove self-excited oscillation.

\emph{Open items.} Robust redesign, rotor-cap and allocation-aware studies are reserved for a companion paper. The shared-function Simulink replication checks implementation consistency only. The PX4 SITL check (Section~\ref{sec:sitl}) runs the autopilot software and its estimator but uses the same plant model; HIL and flight validation remain future work. Behind the estimator, the comparison constrains the roll/pitch-cascade and yaw increments but not the altitude class. A ceiling for that class would address the hover excess of the AA-E designs (12--20\% in $J$ over the original and rerun campaigns); this is a next step, not a result of this paper.

\section{Conclusion}\label{sec:conclusion}
This paper showed that the input supplied to the extended-state observer is a design variable in sampled cascade LADRC of quadrotors. Actuator-model inputs for altitude and attitude and measured-attitude inputs for horizontal motion remove the nominal structural observer-input residual under matched lag and delay; the improvement comes from the rotor-lag model and the measured-attitude input, while measurement-time alignment matters little at the nominal 5-ms delay. With commanded inputs, the searched designs met the requirements only at the lowest observer/controller ratios, with 9.2 times the RMSE of AA-Budget. A full-cascade covariance model predicts the noise-induced inner-loop increment in noise-only hover with white measurement noise within 0.3\%, which turns the outer-to-inner noise path into an explicit tuning constraint. In paired set-point and lemniscate trials, AA-Budget and AA-HalfGain reach 19.2/16.9~mm and 22.7/20.4~mm versus 65.5 and 276.5~mm for PID-Budget, and 29.9 and 26.4~mm versus 72.3~mm under a common strict cascade ceiling, at which the PID uses roughly double their rotor-command increment. The nominal advantage does not extend to large rotor-lag or delay mismatch, where the AA designs are less tolerant than PID.

PX4 software-in-the-loop runs reproduce these results within 2~mm in two independent campaigns. Behind EKF2, a shaping-filter model of the estimation errors predicts the roll/pitch and yaw increments within 5\% and the total rotor-command increment within 14\%, but the tracking variance only approximately. After redesign for this stack under the same specification, AA-Budget-E reduces the set-point and lemniscate RMSEs from 59.1 to 44.0~mm and from 335.8 to 49.5~mm relative to PID-E. It has a lower rotor-command increment in every paired tracking run (S6 mean: 45\% lower before and 15\% after output quantization). AA-HalfGain-E is worse on the set-point scenario. The evidence is limited to simulation and SITL with one plant model and one estimator configuration; robust redesign over an uncertainty set, a ceiling for the altitude class and HIL or flight validation remain future work.

\appendices
\section{Scenarios and reproducibility}\label{app:scenarios}
All trials start at zero position, Euler angles and rates, with rotors at hover; noise and gyro bias are active from startup, and mean wind and turbulence start at 5~s. In S6 (40~s, window 10--40~s) the reference is the step $(x,y,z,\psi)=(0.5,0.5,2,0.5)$ m/rad at $t=0$, so the initial transient is excluded from RMSE but can affect the subsequent trajectory. In S5 (45~s, window 5--45~s) the reference climbs smoothly to 2~m in 5~s and then follows a horizontal lemniscate with 1-m amplitude and 20-s period, with $\psi_d=0$. Both scenarios add the torque disturbance $[0.005\sin(2\pi0.5t),0.005,0]^\top$ N\,m from 15~s. MEX-accelerated runs are checked against full MATLAB trajectories before use, and all figures and table entries are generated from the saved results.

\emph{Seeds.} Nominal, ablation, strict and budget-curve trials: 91001--91030 (SITL S5 and ablations: 91001--91010); mismatch: 92001--92010; noise and latency stress: 93001--93010; 60-s hover: 96001--96010 (fit 96001--96005); 300-s hover validation: 97001--97010 (fit 97001--97005).

\emph{PX4 configuration.} PX4 v1.16.1 SITL in lockstep; IMU integration and gyroscope rate 1~kHz (\texttt{IMU\_INTEG\_RATE}, \texttt{IMU\_GYRO\_RATEMAX}); EKF2 fuses the IMU and external vision for position, velocity and yaw (\texttt{EKF2\_EV\_CTRL}~=~15, \texttt{EKF2\_HGT\_REF}~=~3, \texttt{EKF2\_EV\_DELAY}~=~5~ms, \texttt{EKF2\_EVP\_NOISE}, \texttt{EKF2\_EVV\_NOISE}, \texttt{EKF2\_EVA\_NOISE}~=~0.01~m, 0.02~m/s, 0.05~rad); GPS, barometer, magnetometer and range sensor are not fused; all other parameters, including the 40-Hz gyroscope filter, are PX4 defaults. Delays: 5-ms measurement and external-vision latency; models of the -E designs use a 6-ms delay on the three body rates and none on the other signals.

\emph{Gains, models and scripts.} Table~\ref{tab:gains2} lists the gains of the strict and estimator-aware designs. The budget-curve gains, all full-precision gains, the noise models, the frozen ceilings, the SITL run plans and the scripts that generate each table and figure from the saved runs are stored with the code, together with a manifest of result files and checksums and a step-by-step re-run procedure. Code, controller gains, random seeds, and per-seed results are publicly available at: {\url{https://github.com/Quang61123/actuator-aware-cascade-adrc}}.

\begin{table}[t]\centering\scriptsize
\caption{Gains of the Strict and Estimator-Aware Designs}\label{tab:gains2}
\setlength{\tabcolsep}{2pt}
\textbf{(a) PID}\\[1pt]
\begin{tabular}{llrrrrr}\toprule
Design & Class & $K_p$ & $K_i$ & $K_d$ & $\omega_f$ & $b$\\\midrule
PID-Budget-S & $xy$ & 5.151 & 4.083 & 3.032 & 795.5 & 0.300\\
 & $z$ & 19.462 & 35.684 & 6.876 & 19.985 & 0.0017\\
 & $\phi\theta$ & 71.423 & 58.449 & 9.485 & 301.57 & 0.789\\
 & $\psi$ & 32.215 & 73.392 & 9.064 & 33.064 & 0.0018\\\midrule
PID-E & $xy$ & 5.232 & 4.885 & 2.912 & 15054 & 0.0014\\
 & $\phi\theta$ & 118.261 & 112.004 & 16.746 & 671.71 & 0.983\\
 & $\psi$ & 32.216 & 74.970 & 8.887 & 33.976 & 0.0012\\
\bottomrule\end{tabular}\\[3pt]
\textbf{(b) LADRC}\\[1pt]
\begin{tabular}{lrrrr}\toprule
Design & $xy$ & $z$ & $\phi\theta$ & $\psi$\\\midrule
AA-Budget-S & (1.646,30.811) & (3.029,34.773) & (8.340,156.108) & (4.025,59.598)\\
AA-HalfGain-S & (1.723,39.167) & (3.053,45.203) & (8.726,146.738) & (4.085,76.450)\\
AA-Budget-E & (1.372,31.188) & (3.786,43.466) & (6.808,54.612) & (5.191,9.454)\\
AA-HalfGain-E & (1.137,9.786) & (3.816,56.503) & (5.685,89.187) & (3.125,18.245)\\
\bottomrule\end{tabular}\par\smallskip\begin{minipage}{\linewidth}\footnotesize Notation as in Table~\ref{tab:gains}. PID-E keeps the altitude gains of PID-Budget; AA-Budget-E and AA-HalfGain-E keep the AA altitude bandwidths.\end{minipage}
\end{table}

\section{Search and feasibility-repair details}\label{app:search}
\emph{Sensitivity grid.} During the design search, $M_s$ is evaluated on 600 logarithmically spaced positive frequencies from $10^{-2}$ rad/s to $\pi/T_s$, plus dc, followed by local peak refinement; final candidates are re-evaluated on 4000 positive frequencies plus dc with refined peak searches. The tuning limit 1.98 provides numerical headroom for the final audit at 2. Grid peaks no greater than $1+10^{-8}$ are omitted from local refinement to avoid round-off extrema on the unity plateau; they remain included in the grid maximum.

\emph{Bandwidth search.} For each observer-to-controller ratio $r=\omega_o/\omega_c$ on a prescribed grid, a bandwidth scan identifies feasible intervals of \eqref{eq:opt}, whose boundaries are then refined. The feasible design with the smallest predicted tracking variance is retained and rechecked with the full-resolution overshoot calculation and the 4000-point sensitivity search. The inner and altitude searches use an overshoot screen of 9.95\% (horizontal reselection: 10\%), and all final designs are audited against 10\%.

\emph{Cascade thresholds.} A 1\% feasibility margin is applied below the selected cascade threshold in every call to the horizontal reselection routine, including strict and budget-curve PID candidates.

\emph{Strict designs and repair.} The inner LADRC bandwidths are initially scaled by 0.8, and the PID inner-loop bandwidth by 0.9 with its controller shape preserved. If necessary, a full-resolution feasibility-repair step further reduces controller or observer bandwidths, or the PID set-point weight, until the constraints are satisfied. The horizontal gains are then redesigned with the same local activity limits and the selected cascade constraint. Because the procedure combines finite grid searches, boundary refinement and feasibility repair, the reported parameters are the best feasible designs found, not global optima.

\emph{Redesigns under the response requirement.} PID-Budget-S was redesigned by multistart sequential quadratic programming (SQP; 20 starts for each of $z$ and $\psi$ and 20 for a joint $\phi\theta$+$xy$ search under the strict cascade ceiling) with all constraints of \eqref{eq:opt}; this search additionally required a spectral radius of at most 0.9999, tighter than the specification $\rho<1$. The PID budget-curve points $\gamma=0.5$ and 0.75 used the same multistart search. Infeasible results then went through a general repair step that reduces the controller bandwidth by factors of 0.985 and then 0.9 and lowers the set-point weight when the overshoot is exceeded. The levels are processed in increasing $\gamma$. The feasible design of the smaller $\gamma$ is included as a start and as a candidate at the larger $\gamma$, where it remains feasible because only the local ceilings change. Because three constraints are active at once, these searches use an overshoot target of 9.99\%, while the final audit uses 10\%. A1-Commanded and RO-HalfGain were searched for every class on 24 ratios $\omega_o/\omega_c\in[1,60]$ (the 20 ratios from 1.5 to 60 plus 1.0--1.36) with 60 controller bandwidths per ratio. For the -E designs, PID-E used multistart SQP (20 starts for $\psi$ and for the joint $\phi\theta$+$xy$ search). Its $xy$ set-point weight was then lowered in 18 steps of 0.1\% to move an overshoot of $10+10^{-6}$\% inside the limit ($b$ affects only the reference response). AA-Budget-E and AA-HalfGain-E used a $20\times40$ grid for $\psi$ and a $12\times12$ grid for $\phi\theta$, each feasible attitude point followed by the horizontal search.

\emph{Noise-model fitting.} The estimator-aware models of Section~\ref{sec:sitl} are vector autoregressive models of the class error vectors, fitted by least squares for each channel class and design on the hover seeds 96001--96005 and written in the innovations form \eqref{eq:shaping}; the order (at most 15) minimizes the Bayesian information criterion among the stationary fits. Vector autoregressive models were preferred to a subspace (n4sid) fit, which missed the increment variances on the training data by factors of 30 to 9000. Only stationary fits are admitted, because a nonstationary fit gives an unbounded prediction; this requirement was fixed before any comparison with the SITL hover data and affected a single yaw model. No modelling choice, including the signal delays, was changed after the predictions were compared with the 60-s SITL hover runs. Propagating the measured cross-spectral densities instead of the fitted filters changes the predicted increments by less than 2\%. A pooled model is fitted by least squares on the concatenated training segments of all designs in the pool, each segment demeaned and every output scaled by its pooled standard deviation; the -E designs were selected with the model pooled over the six designs of the 60-s logs. For the three-level validation of Table~\ref{tab:estaware} the same procedure was applied to the 300-s logs, per design (level a) and pooled over the nine designs (level b). The design-stage model had no delay on the yaw-rate error; for level (c) it was refitted on the same 60-s logs with the 6-ms convention on all body rates, which changes the yaw predictions by less than 0.5\%, and all constraints of the -E designs remain satisfied with it.

\section{Verification and lockstep audit}\label{app:audit}
\emph{MATLAB checks.} An independent augmented-state increment expression agrees with the direct Lyapunov formula for all four nominal candidates and channel classes (maximum relative difference 2.08e-14). Halving the RK4 plant substep changes nominal RMSE by at most 9.63e-08 mm, and a shared-function Simulink replication of all nominal candidates on S6 gives a maximum state discrepancy of 0. The rerun of Section~\ref{sec:numver} covers the 1830 primary realizations and the 1480 supplemental simulation calls of A1-Commanded, RO-HalfGain, PID-Budget-S, the PID points of Fig.~\ref{fig:budget}, the -E designs and the PID set-point weights. The hover measurements in Table~\ref{tab:cascadenoise} use ten paired 60-s noise-only trials per design (seeds 96001--96010, window 20--60 s, no wind, bias or external disturbance, no rotor clipping).

\emph{PX4 SITL rerun.} All 1288 planned SITL runs were executed in the rerun (30 of them belong to a design iteration that was not needed and are not analysed). The divergence classification of every run with a counterpart in the reported campaign is unchanged (six divergences, all in the mismatch set at a true rotor lag of 120~ms). The S6 reductions of $J$ and $J_q$ relative to PID-E are 44.6 and 14.5\% (reported campaign: 45 and 15\%).

\emph{Skipped lockstep steps.} Table~\ref{tab:miss} lists the runs in which PX4 skipped the output of a step. The rule, declared before the sensitivity analysis, was to rerun a comparison group only if removing its pairs with skipped steps changed a paired conclusion. This did not occur, so no run was repeated. AA-Budget-E minus PID-E in S6 is $-14.6\pm1.2$~mm over the 23 pairs without skipped steps versus $-15.0\pm1.1$~mm over all 30; AA-Budget-S minus PID-Budget-S is $-41.9\pm2.4$~mm over 21 pairs; and AA-Budget-E minus PID-Budget in S5 is $-226.0\pm14.8$~mm over the three pairs without skipped steps versus $-229.0\pm6.3$~mm over all ten. Part of the hover runs and nine S6 runs of PID-Budget-S overlapped with a parallel computation on the host, which increased the skipped steps; no S6 or S5 run of the -E designs did.

\begin{table}[t]\centering\scriptsize
\caption{Skipped Lockstep Steps in PX4 SITL}\label{tab:miss}
\setlength{\tabcolsep}{2.5pt}\begin{tabular}{lrrrr}\toprule
Group & Runs & Skip & Steps (max) & Par.\\\midrule
Hover 60 s, -E designs (refit) & 30 & 7 & 9 (3) & 24\\
S6, -E designs & 90 & 10 & 11 (2) & 0\\
S5, -E designs & 30 & 2 & 2 (1) & 0\\
Hover 300 s, -E designs & 30 & 11 & 16 (3) & 0\\
Hover 300 s, primary and strict & 60 & 35 & 68 (9) & 3\\
S6, PID-Budget-S & 30 & 9 & 22 (6) & 9\\
Hover 60 s, PID-Budget-S & 10 & 1 & 1 (1) & 2\\
S6, A1, ratio $\ge1.5$ (sensitivity) & 10 & 5 & 30 (9) & 5\\
S6, A1, ratio $\ge1$ & 10 & 1 & 1 (1) & n.r.\\
S6, RO-HalfGain & 30 & 3 & 3 (1) & n.r.\\
S5, RO-HalfGain & 10 & 1 & 1 (1) & n.r.\\
S6, PID-Budget (set-point rule) & 30 & 8 & 9 (2) & n.r.\\
S5, PID-Budget (set-point rule) & 10 & 6 & 6 (1) & n.r.\\
S6 bypass, PID-Budget (set-point rule) & 10 & 2 & 2 (1) & n.r.\\\midrule
Total & 390 & 101 & 181 & \\
\bottomrule\end{tabular}\\[2pt]
\footnotesize Skip: runs with at least one skipped step; steps: total (maximum per run); par.: runs overlapping a parallel host job.  n.r.: not recorded; only light post-processing ran on the host during these runs. In the independent rerun (Section~\ref{sec:numver}), 76 of these 390 runs had skipped steps (1088 steps in total, 999 of them in a single 300-s hover run of AA-HalfGain-S); no rerun result enters this table or the noise models, and overlap with host jobs was not recorded for the rerun.
\end{table}


\bibliographystyle{IEEEtran}
\bibliography{references}

\end{document}